\documentclass[10pt,twocolumn,letterpaper]{article}

\usepackage[pagenumbers]{cvpr} 

\usepackage{graphicx}
\usepackage{amsmath}
\usepackage{amssymb}
\usepackage[dvipsnames]{xcolor}
\usepackage{booktabs}
\usepackage{xcolor}
\usepackage{soul}
\usepackage{pifont}
\usepackage{fontawesome}
\usepackage[accsupp]{axessibility} 

\usepackage[pagebackref,breaklinks,colorlinks]{hyperref}
\hypersetup{
pdftitle={EVREAL: Towards a Comprehensive Benchmark and Analysis Suite for Event-based Video Reconstruction},
pdfsubject={Computer Vision, Event Vision},
pdfauthor={Burak Ercan, Onur Eker, Aykut Erdem, Erkut Erdem},
pdfkeywords={Event cameras, Dynamic Vision Sensor, Event-based Vision, Video Reconstruction}
}
\usepackage{enumitem}

\usepackage[capitalize]{cleveref}
\crefname{section}{Sec.}{Secs.}
\Crefname{section}{Section}{Sections}
\Crefname{table}{Table}{Tables}
\crefname{table}{Tab.}{Tabs.}

\def\cvprPaperID{19} %
\def\confName{CVPR}
\def\confYear{2023}

\newcommand{\XYZ}[1]{\left\{0,\ldots,{#1}-1\right\}}

\definecolor{somegray}{rgb}{0.5, 0.5, 0.5}
\newcommand{\darkgrayed}[1]{\textcolor{somegray}{#1}}
\makeatletter
\newcommand*\titleheader[1]{\gdef\@titleheader{#1}}
\AtBeginDocument{%
  \let\st@red@title\@title
  \def\@title{%
    \vskip-4.5em
    \bgroup\normalfont\large\centering\@titleheader\par\egroup
    \vskip1.5em\st@red@title}
}

\makeatother

\titleheader{\darkgrayed{This paper has been accepted for publication at the IEEE Conference on\\
Computer Vision and Pattern Recognition Workshops (CVPRW), Vancouver, 2023. \\
\copyright IEEE}}

\title{EVREAL: Towards a Comprehensive Benchmark and Analysis Suite for Event-based Video Reconstruction}

\begin{document}

\author{Burak Ercan \textsuperscript{1,2}
\quad\quad Onur Eker \textsuperscript{1,2}
\quad\quad Aykut Erdem \textsuperscript{3,4}
\quad\quad Erkut Erdem \textsuperscript{1,4} \vspace{0.2cm} \\
\textsuperscript{1} Hacettepe University, Computer Engineering Department ~~ \textsuperscript{2} HAVELSAN Inc.  \\
\textsuperscript{3} Ko\c{c} University, Computer Engineering Department ~~ \textsuperscript{4} Ko\c{c} University, KUIS AI Center \\
}
\maketitle

\begin{abstract}
Event cameras are a new type of vision sensor that incorporates asynchronous and independent pixels, offering advantages over traditional frame-based cameras such as high dynamic range and minimal motion blur. However, their output is not easily understandable by humans, making the reconstruction of intensity images from event streams a fundamental task in event-based vision. While recent deep learning-based methods have shown promise in video reconstruction from events, this problem is not completely solved yet. To facilitate comparison between different approaches, standardized evaluation protocols and diverse test datasets are essential. This paper proposes a unified evaluation methodology and introduces an open-source framework called EVREAL to comprehensively benchmark and analyze various event-based video reconstruction methods from the literature. Using EVREAL, we give a detailed analysis of the state-of-the-art methods for event-based video reconstruction, and provide valuable insights into the performance of these methods under varying settings, challenging scenarios, and downstream tasks. 
\end{abstract}

\vspace{-0.15cm}
More results, analyses, and source code available at:~ \\
{\small\url{https://ercanburak.github.io/evreal.html}}
\vspace{-0.05cm}

\section{Introduction}
\label{sec:intro}
Event cameras are a new type of biologically-inspired vision sensor that have the potential to overcome the limitations of conventional frame-based cameras. Unlike traditional cameras, event cameras have pixels that work independently and asynchronously from one another. Each pixel detects changes in the relative brightness of its local area and generates an event signal when the change exceeds a certain threshold. Hence, the output is a continuous stream of events, with each event containing information about the location, the polarity of the intensity change, and the precise time of the detected change. The rate of event generation varies depending on the scene characteristics, with more events being triggered for scenes showing rapid motion or instant changes in brightness and texture. Due to their unique working principles, event cameras provide many advantages, such as high dynamic range, high temporal resolution, and minimal motion blur~\cite{gallego2020event}.

While event streams have many desirable properties, they have one major disadvantage -- humans cannot easily understand event streams in the same way as intensity images. Hence, a fundamental task in event-based vision literature is reconstructing intensity images from event streams. Reconstructing high-quality videos from events also allows for employing existing frame-based computer vision methods developed for several downstream tasks to event data in a straightforward manner \cite{rebecq2019events}.

While deep learning-based methods have made impressive progress in reconstructing videos from event streams lately (e.g., \cite{rebecq2019high,stoffregen2020reducing,weng2021event}), this research problem is still not completely solved. This can be primarily attributed to the event representations being used in these state-of-the-art approaches, which can cause some latency issues. Moreover, training these methods relies heavily on synthetically created datasets. Consequently, these methods may produce suboptimal video reconstructions that suffer from issues such as blur, low contrast, or smearing artifacts.

A significant effort has been put forth to find better ways to evaluate event-based video reconstruction methods and assess the visual qualities of reconstructed videos. There are several distinct evaluation methodologies involving different datasets, event representations, post-processing steps, quantitative metrics, and downstream tasks (see \cref{tab:features} for an overview, and refer to the supplementary materials for a more detailed discussion). However, the lack of a standard evaluation procedure makes it hard to fairly compare the performances of different methods. The details of the evaluation procedures are sometimes not clearly defined, even though each minor detail may significantly alter the results. This also poses challenges for other researchers to reproduce the reported results. This motivates the need for open-source codes and standardized protocols for evaluation.

Comparing different methods requires not only well-defined evaluation protocols but also a diverse set of test datasets that cover various real-world settings. Large-scale benchmarks have been instrumental in advancing many computer vision tasks, as demonstrated by \mbox{ImageNet~\cite{russakovsky2015imagenet}} for image classification and \mbox{MS-COCO~\cite{lin2014microsoft}} for object detection, providing results that are generalizable to unseen real-world data. However, since event-based vision is a relatively new field compared to classical frame-based computer vision, the current datasets used for assessing event-based video reconstruction are limited in scale and scope, confined to specific domains, scenes, camera types, and motion patterns. To ensure the generalizability of the results and evaluate the methods' effectiveness in more real-world scenarios, it is essential to assess their performances on a large variety of datasets showing different characteristics.

\begin{table*}[!t]
  \begin{center}
  \small
  \setlength{\tabcolsep}{5pt}
  \begin{tabular}{lccccccccccc}
    \toprule
    \makebox[0pt][l]{Evaluation} & Test & \# of & Compared & Metrics  & Comp. & Chlng. & Downst.  & Robust. & Open \\ 
    Setup in & Datasets & Frames & Methods & (FR/NR) & Eff. & Scnrs. & Tasks & Exp. & Source \\
    \midrule
    \cite{wang2019event_cvpr}     & \cite{bardow2016simultaneous} & 0.2K & \cite{bardow2016simultaneous, munda2018real, wang2019event_cvpr} & --/B &  & \faImage &                        &  &  \\ 
    \cite{rebecq2019high}         & \cite{mueggler2017event} & ~1.9K  & \cite{bardow2016simultaneous, munda2018real, rebecq2019high} & MSLT/-- & \checkmark & \faImage & \faImage~\faImage~\faBarChart~\faBarChart & \faImage \\ 
    \cite{Scheerlinck20wacv}      & \cite{mueggler2017event} & ~1.9K  & \cite{bardow2016simultaneous, munda2018real, rebecq2019high, Scheerlinck20wacv} & MSL/-- & \checkmark & \faImage &                        &  &  \\ 
    \cite{stoffregen2020reducing} & \cite{mueggler2017event, zhu2018multivehicle, stoffregen2020reducing} & 28.7K & \cite{rebecq2019high, Scheerlinck20wacv, stoffregen2020reducing} & MSL/-- &      &  &                        &  &  \\ 
    \cite{cadena2021spade}        & \cite{mueggler2017event} & ~3.1K  & \cite{rebecq2019high, Scheerlinck20wacv, cadena2021spade} & MSLT/R & \checkmark &  &    \faBarChart                     &  & \checkmark \\ 
    \cite{weng2021event}          & \cite{mueggler2017event, zhu2018multivehicle, stoffregen2020reducing} & ~28.7K & \cite{rebecq2019high, Scheerlinck20wacv, stoffregen2020reducing, weng2021event} & MSL/-- &  & \faImage &                        &  & \checkmark \\ 
    \cite{paredes2021back}        & \cite{mueggler2017event, stoffregen2020reducing} & ~17.4K & \cite{rebecq2019high, Scheerlinck20wacv, stoffregen2020reducing, paredes2021back} & MSL/-- &      & \faImage &                        &  &  \\ 
    \cite{zhu2022event}           & \cite{mueggler2017event, zhu2018multivehicle, stoffregen2020reducing} & ~28.7K & \cite{rebecq2019high, Scheerlinck20wacv, stoffregen2020reducing, cadena2021spade, zhu2022event} & MSL/-- & \checkmark &  &                        & \faBarChart &  \\ 
    \cite{zhang2022formulating}   & \cite{mueggler2017event, orchard2015converting} & 1.9K  & \cite{rebecq2019high, stoffregen2020reducing, paredes2021back, zhang2022formulating} & MSL/-- & \checkmark & \faImage &                        & \faBarChart &  \\ 
    Ours                          & \cite{mueggler2017event, zhu2018multivehicle, rebecq2019high, stoffregen2020reducing, tulyakov2022time} & ~47.7K & \cite{rebecq2019high, Scheerlinck20wacv, stoffregen2020reducing, cadena2021spade, paredes2021back, weng2021event} & MSL/BNM & \checkmark & \faBarChart  & \faBarChart ~\faBarChart~\faBarChart       & \faBarChart & \checkmark \\ 
    \bottomrule
  \end{tabular}
  \end{center}
  \vspace{-0.5cm}
  \caption{\textbf{A summary of experimental setups considered in earlier work.} We provide a comparison of our proposed EVREAL framework to the experimental evaluation setups reported in the existing work in terms of  datasets being used, methods compared, number of reconstructed frames used in quantitative analysis, and metrics being utilized. We also indicate whether each evaluation setup includes analysis of computational efficiency, challenging scenarios (fast motion, low light, or high-dynamic range), downstream tasks, and robustness. Finally, we mark whether the implementation of this evaluation setup is open-sourced or not. In the metrics column, FR and NR stand for full-reference and no-reference metrics, respectively. M:MSE, S:SSIM~\cite{wang2004image}, L:LPIPS~\cite{zhang2018unreasonable}, and T:Temporal Consistency~\cite{lai2018learning} are the full-reference metrics, while R:RMS contrast, B:BRISQUE~\cite{mittal2012no}, N:NIQE~\cite{mittal2012making}, and M:MANIQA~\cite{yang2022maniqa} are the no-reference metrics. In the Challenging Scenarios, the Downstream Tasks and Robustness Experiments columns, each \faImage~symbol denotes a reported qualitative analysis and a \faBarChart~symbol represents a quantitative analysis being performed along with a qualitative comparison. \vspace{-0.5cm}}
  \label{tab:features}
\end{table*}

Event data is handy in scenarios where traditional frame cameras fail, such as scenes captured under low-light conditions or with rapid motion, and underexposed or overexposed regions. Hence, it is of utmost importance to evaluate the effectiveness of event-based video reconstruction models in those challenging situations. However, as traditional frame-based cameras especially struggle in these scenarios, collecting high-quality reference frames is a challenging task on its own. This paradox makes it difficult to quantify the success of event-based video reconstruction methods where they are most needed.

Even in scenarios where it is possible to collect high-quality reference frames with minimal motion blur and optimal exposure, assessing image quality remains a subjective endeavor. Hence, the current studies generally consider a perceptual metric like learned perceptual image patch similarity (LPIPS)~\cite{zhang2018unreasonable} along with distortion-aware metrics like PSNR and structural similarity (SSIM)\cite{wang2004image}. However, as a full-reference metric, LPIPS is trained on distortions that are not commonly seen in the reconstructed intensity images from event data. Hence, this raises some doubts about the significance of these perceptual comparisons.

Reconstructing images from events is a complex task. It depends on many variables that can affect the performance of the methods. These include sensor noise characteristics, sensor parameters, event generation rate, event grouping scheme, grouping rate, frame reconstruction rate, and temporal regularity. Despite their importance, the literature often overlooks the robustness of the methods to changes in these variables. Therefore, it is crucial to evaluate the sensitivity of the methods to these variables and to consider their performance under changing conditions. A method that performs well under specific settings may not be suitable for general use when these variables are expected to change.

Event cameras are known for their low-latency and non-redundant data flow, making them ideal for scenarios that require real-time and low-power processing. As a result, the computational efficiency of event-based video reconstruction methods is just as important as the visual quality of reconstructions. Neglecting this aspect in a benchmark could lead to choosing a method that provides high-quality reconstructions, but is impractical for real-time processing.

To address these issues and facilitate progress in event-based video reconstruction, in this study, we propose EVREAL, Event-based Video Reconstruction Evaluation and Analysis Library, an open-source framework based on PyTorch~\cite{paszke2019pytorch}. Our framework offers a unified evaluation pipeline to benchmark pre-trained neural networks and a result analysis tool to visualize and compare reconstructions and their scores. We use a large set of real-world test sequences and various full-, and no-reference image quality metrics to perform qualitative and quantitative analysis under diverse conditions, including challenging scenarios such as rapid motion, low light, and high dynamic range, many of which have not been reported before. Moreover, we conduct experiments to assess the performance of each method under variable conditions and analyze their robustness to these varying settings. We also evaluate the quality of video reconstructions via downstream tasks like camera calibration, image classification, and object detection. This extrinsic evaluation can be considered a proxy metric for image quality and a task-specific metric if the goal of event-based video reconstruction is to perform these downstream tasks.

In~\Cref{tab:features}, we present an overview of our experimental setup in comparison to prior work. Along with this paper, we build a website to share our results and findings, together with the source code to reproduce them
. We also intend to update this webpage on a regular basis as new event-based video reconstruction methods are proposed. Our contributions in this paper can be summarized as follows:
\begin{itemize}[noitemsep,topsep=0pt,leftmargin=*]
\item We propose a unified evaluation methodology and an open-source framework to benchmark and analyze event-based video reconstruction methods from the literature. 
\item Our benchmark includes additional datasets, metrics, and analysis settings that have not been reported before. We present quantitative results on challenging scenarios involving rapid motion, low light, and high dynamic range.
\item Moreover, we conduct additional experiments to analyze the robustness of methods under varying settings such as event rate, event tensor sparsity, reconstruction rate, and temporal irregularity.
\item To further examine the quality of the reconstructions, we provide quantitative analysis on several downstream tasks, including camera calibration, image classification, and object detection.
\end{itemize}

\section{Methodology of Evaluation and Analysis}
\label{sec:method}

\subsection{Task Description}

Suppose we have a stream of events $\{e_i\}$ containing $N_E$ events and spanning $T$ sec. Each event $e_i=(x_i,y_i,t_i,p_i)$ in the stream represents a change in brightness perceived by the sensor, and contains information about the location $(x_i, y_i)$, the timestamp $t_i$, and the polarity $p_i$ of this intensity change. Here, $t_i \in [0, T]$, $p_i \in \{+1,-1\}$, $x_i \in \XYZ{W}$, and $y_i \in \XYZ{H} $ for all $ i \in \XYZ{N_E}$, with $W$ and $H$ denoting the width and height of the sensor array, respectively. Given these events, the task aims to generate a stream of $N_I$ images~$\{ \hat{I}_k \}$, corresponding to the same $T$ sec. period as the events. Each image $\hat{I}_k$ represents the absolute brightness of the scene, as if it were captured by a standard frame-based camera at a particular time $s_k$ within the time period of $T$ seconds, where $k$ ranges from $1$ to $N_I$, and $\hat{I}_k$ ranges from $0$ to $1$.

\begin{figure*}[t]
\centering
\includegraphics[width=0.9\linewidth]{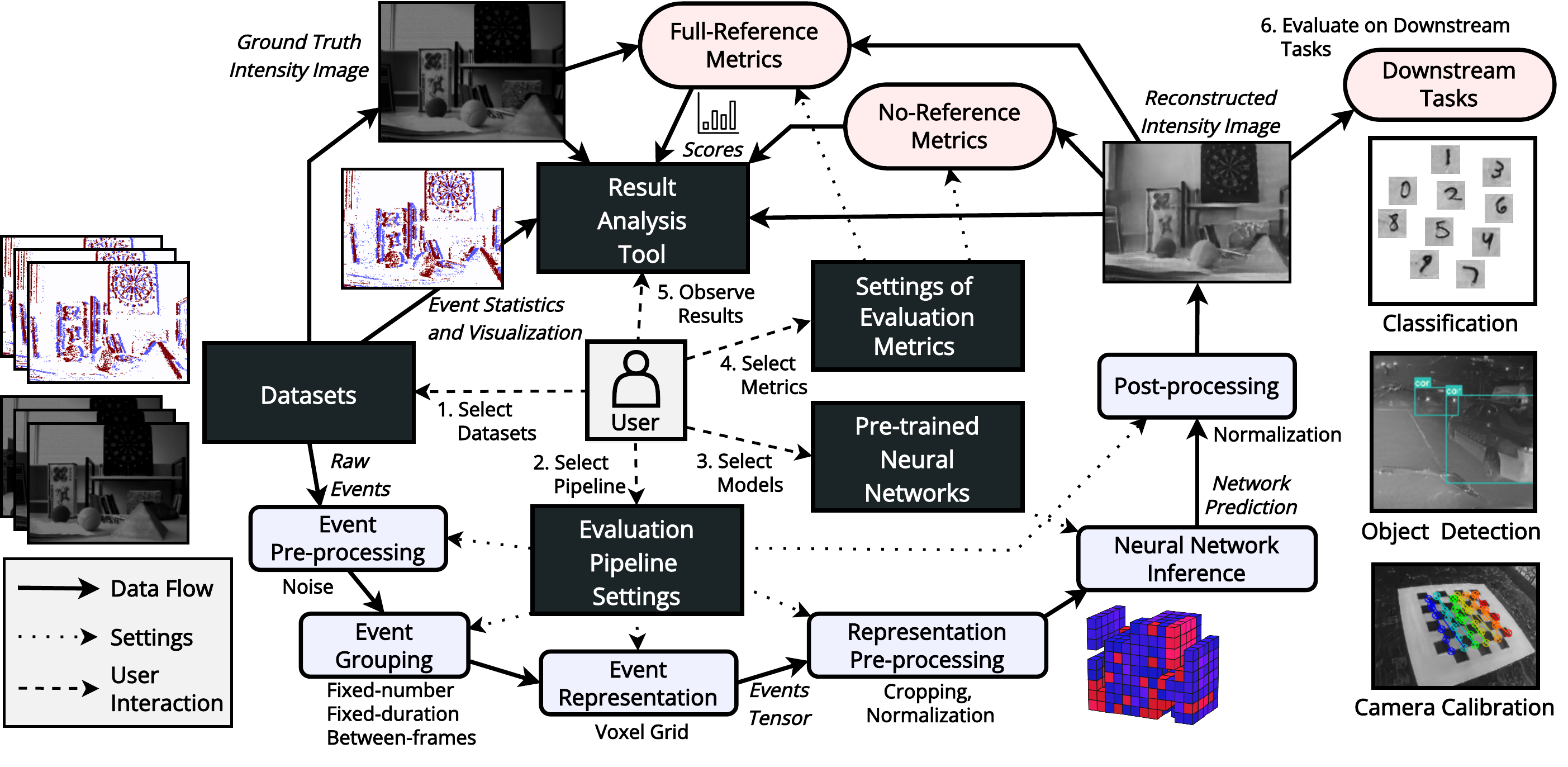}
\caption{\textbf{An overall look at our proposed EVREAL (Event-based Video Reconstruction -- Evaluation and Analysis Library) toolkit.}\vspace{-0.4cm}}
\label{fig:evreal}
\end{figure*}

\subsection{Evaluation Framework and Pipeline}
EVREAL implements several standardized components crucial for deep event-based video reconstruction models, including \emph{event pre-processing}, \emph{event grouping}, \emph{event representation}, \emph{representation processing}, and \emph{image post-processing} (see \cref{fig:evreal}). We have included components to evaluate the visual quality of each frame in the generated videos, which are split into \emph{full-reference} metrics and \emph{no-reference} metrics. The former is utilized when high-quality, distortion-free ground truth frames are available, whereas the latter are used when ground truth frames are of low quality or not available at all (refer to \cref{sec:metrics}). 
EVREAL also includes an analysis tool. Given a set of reconstructions generated by one or more methods, it collects ground truth frames, event visualizations, event rate statistics, and instantaneous values for a set of quantitative metrics. It then generates an output video that displays this data in a time-synchronized manner, including plots of quantitative metrics (see the supplementary for a sample output video). Our tool is particularly valuable in pinpointing specific limitations and failure cases of methods. For instance, it can reveal situations where noisy reconstructions significantly impact future reconstructions due to the sequential nature of the method. Such scenarios can be visually identified from the plots of quantitative metrics.
To assess the practical use of a given method, our framework allows for evaluating it on several downstream tasks. Specifically, we analyze the performance of tested models on three downstream tasks, \emph{object detection}, \emph{image classification}, and \emph{camera calibration} (\cref{sec:downstream}).
We want to emphasize that our objective with this work is to conduct a comprehensive evaluation and analysis of existing pre-trained models from the literature in order to characterize them rather than to provide a ranking of them. Developing a new model is also beyond the scope of this work. In the following, we provide detailed descriptions of the components of our evaluation framework.

\vspace{0.05cm}
\noindent \textbf{Event pre-processing.} This component can be employed to process raw events before grouping them. Possible pre-processing operations include event temporal downsampling and adding artificial event noise, to perform robustness experiments under these conditions.

\vspace{0.05cm}
\noindent \textbf{Event grouping.} Each event in isolation contains limited information about the scene, so a common practice is to group a number of events together and process them as a whole. We consider the following grouping options:
\begin{itemize}[noitemsep,topsep=0pt,leftmargin=*]
    \item \textbf{Fixed-number:} We group every $N_G$ number of events such that the $k$th event group can be defined as:
    \begin{equation}
    \label{eq:group_fixed_events}
    G_k \doteq \{e_i \mid kN_G \leq i < (k+1)N_G \}
    \end{equation}
    Here, the rate at which the groups are formed varies according to the incoming event rate.
   
    \item \textbf{Fixed-duration:} We group events according to non-overlapping time windows with a fixed duration of $T_G$ secs. The $k$th event group contains all events with timestamps $t_i$ falling within the $k$th time window, defined as:
    \begin{equation}
    \label{eq:group_fixed_time}
    G_k \doteq \{e_i \mid kT_G \leq t_i < (k+1)T_G \}
    \end{equation}
    In this scheme, the number of events in each group varies according to the incoming event rate.
    
    \item \textbf{Between-frames:} If we also have ground truth intensity frames together with the incoming event stream, we can group events such that every event between consecutive frames belongs to the same group. Therefore, the set of events in the $k$th event group can be defined as follows:
    \begin{equation}
    \label{eq:group_between_frames}
    G_k \doteq \{e_i \mid s_k \leq t_i < s_{k+1} \}
    \end{equation}
    If the ground truth frames arrive at a fixed rate, then this option is a special case of the fixed temporal window grouping. Note that, however, the time difference between consecutive frames may not be fixed all the time, \textit{e.g.}, due to changing exposure times of the camera in real-world datasets, or due to adaptive rendering schemes of simulators in synthetic datasets.
\end{itemize}

\vspace{0.05cm}
\noindent \textbf{Event representation.} To utilize deep CNN architectures for event-based data, a common choice is to accumulate grouped events into a grid-structured representation such as a voxel grid~\cite{zihao2018unsupervised}. We also follow this approach in our evaluation procedure. The details of this event representation are given in the supplementary material.

\vspace{0.05cm}
\noindent \textbf{Representation pre-processing.} After forming a representation from grouped events, it is possible to pre-process this representation before feeding it to the neural network, such as cropping or applying normalization. During our experimental analysis, we apply such pre-processing steps when required by the respective method.

\vspace{0.05cm}
\noindent \textbf{Neural network inference.} This module is used for predicting intensity frames given the event representation by employing the pre-trained neural network model chosen by the user. As mentioned earlier, we use PyTorch here.

\vspace{0.05cm}
\noindent \textbf{Post-processing.} It is also possible to post-process the intensity frame that the network predicts, by utilizing procedures like robust min/max normalization, as done in~\cite{rebecq2019high}. While performing our experiments, we also apply any post-processing operations as suggested by each method.

\subsection{Tested Approaches}
We compare seven methods from the literature that have PyTorch based open-source model codes and pre-trained models. These methods include E2VID~\cite{rebecq2019high}, FireNet~\cite{Scheerlinck20wacv}, FireNet+ and E2VID+~\cite{stoffregen2020reducing}, SPADE-E2VID~\cite{cadena2021spade}, SSL-E2VID~\cite{paredes2021back}, and ET-Net~\cite{weng2021event}. Note that E2VID+ and SSL-E2VID share the same deep network architecture as E2VID, while FireNet+ employs the same architecture as FireNet. Here we utilize the pre-trained models shared publicly by the authors and evaluate them on the same datasets under a common experimental evaluation setup. For E2VID, FireNet, and SSL-E2VID, we normalize event voxel grids as suggested in these methods. Furthermore, we perform robust min/max normalization as a post-processing step for E2VID and SSL-E2VID. For SSL-E2VID, we also apply the exponential function before this min/max normalization.

\subsection{Quantitative Image Quality Metrics} \label{sec:metrics}
To quantitatively assess the quality of videos reconstructed from events, we use both full-reference and no-reference metrics. Full-reference metrics, as their name implies, provide a quality score for an image in regard to a given reference image. In contrast, no-reference metrics do not require any ground truth image and give perceptual quality scores by directly processing input images. 

We utilize three full-reference evaluation metrics: MSE, SSIM~\cite{wang2004image}, and LPIPS~\cite{zhang2018unreasonable}. These metrics are employed only when high-quality, distortion-free ground truth frames are available. While the MSE and SSIM are better suited for capturing distortions, LPIPS measures the perceptual similarity by a deep network trained to conform with human visual perception. Furthermore, we utilize three no-reference metrics: BRISQUE~\cite{mittal2012no}, NIQE~\cite{mittal2012making}, and MANIQA~\cite{yang2022maniqa}. These metrics are used when the ground truth frames are of low quality or when they are not available at all. BRISQUE and NIQE are traditional metrics that employ hand-crafted features and measure conformity to natural scene statistics, considering various synthetic and authentic distortions such as blur, noise, and compression. On the other hand, MANIQA is a deep-learning based method employing vision transformer architecture~\cite{dosovitskiy2020image}, which is trained in an end-to-end manner to assess perceptual image quality while specifically focusing on distortions seen in the outputs of neural network based image restoration algorithms.

The results of the aforementioned metrics can be influenced by specific settings. Hence, to ensure consistency, we provide the detailed settings that we use in the supplementary material. Although performing histogram equalization before calculating image quality metrics (as in \cite{rebecq2019events, Scheerlinck20wacv, paredes2021back}) is supported by EVREAL, we do not perform this operation in our experiments reported in this paper.

\subsection{Datasets}

We adopt commonly used sequences from three datasets in our evaluation framework, namely the Event Camera Dataset (ECD) \cite{mueggler2017event}, the Multi Vehicle Stereo Event Camera (MVSEC) dataset~\cite{zhu2018multivehicle}, the High-Quality Frames (HQF) dataset. In addition, we use handheld sequences from the Beam Splitter Event and RGB (BS-ERGB) Dataset~\cite{tulyakov2022time}. We use full-reference metrics to evaluate the performance of the models on these sequences. To assess the performance of the models in challenging scenarios, we also use no-reference metrics and include sequences with fast motion, low light, and high dynamic range scenes. Specifically, we use the latter parts of ECD sequences where camera movements increase to evaluate fast motion (denoted as ECD-FAST), night driving sequences from the MVSEC to evaluate low light (denoted as MVSEC-NIGHT), and HDR sequences from \cite{rebecq2019high} to evaluate high dynamic range scenes. 
Please refer to the supplementary material for detailed information about these datasets.

\subsection{Robustness Analysis}

To analyze the factors that affect the performance of reconstructing images from events, several variables need to be considered. In this paper, we investigate the impact of four critical ones: \emph{event rate}, \emph{event tensor sparsity}, \emph{image reconstruction rate}, and \emph{temporal irregularity}. To evaluate the results, we utilize the LPIPS metric and employ commonly used sequences from the ECD, MVSEC, and HQF datasets, as mentioned earlier. In the following sections, we provide detailed descriptions of these experiments.

\vspace{0.05cm}\noindent \textbf{Event rate.}
To evaluate the robustness of the methods to varying event rates, we employ between-frames event grouping and collect statistics on event rates, measured in events per second, for each group. We then reconstruct intensity images using each method based on the event groups, and calculate LPIPS scores for each time step. We divide the event rate spectrum into ten equally spaced bins and compute the mean LPIPS scores for each bin and method. This enables us to assess the performance of each method under different event rate conditions and determine which methods are most robust to changes in event rate.

\vspace{0.05cm}
\noindent \textbf{Tensor sparsity.}
To analyze how the sparsity of event tensors affect the performance of each method, we carry out experiments utilizing fixed-number grouping and a tolerance of 1 ms to match the reconstructions with ground truth frames. With this grouping approach, each group contains the same number of events, resulting in event tensors with the same sparsity level. Specifically, we conduct 9 different experiment runs, with event numbers ranging from 5K to 45K. We then compute the mean LPIPS scores for each experiment run and for each method. 

Note that if there exists slow motion or little texture in the scene, using fixed-number grouping can result in event groups that span a large temporal window when the event rate is small. Furthermore, the motion or texture captured by the event camera might be contained in a small region of pixels rather than being homogeneously distributed to all of the sensor area. In that case, the temporal discretization performed in the event representation (to a fixed number of temporal bins) means more compression of the temporal information, and this might result in reconstruction artifacts such as saturation or blur in these regions. The tensor sparsity experiments aid us in assessing each method's robustness to these situations.

\vspace{0.05cm}
\noindent \textbf{Reconstruction rate.}
To evaluate the impact of changing frame reconstruction rates on each method's performance, we conduct experiments using fixed-duration grouping, which generates a fixed number of frames per second. We perform ten experiment runs, each with a different event grouping duration ranging from 10 ms to 100 ms, which correspond to frame reconstruction rates from 10 FPS to 100 FPS. We use a tolerance of 1 ms to match the reconstructions with ground truth frames. We then compute the average LPIPS values for each experiment run and method to determine their performance under different frame rates.

\vspace{0.05cm}
\noindent \textbf{Temporal irregularity.}
To evaluate the robustness of each method in generating frames at irregular intervals, we conduct experiments by removing a certain percentage of ground truth frames from each sequence and by using between-frames event grouping to group events between the remaining frames. In particular, we conduct ten experiment runs with different discarding ratios ranging from 0.0 (standard case) to 0.9. We then calculate the mean LPIPS scores obtained for each experiment run for each method.

\subsection{Analysis on Downstream Tasks} \label{sec:downstream}
Event cameras, due to their unique characteristics, can provide a viable alternative to traditional frame-based cameras in challenging conditions. As a result, using images reconstructed from event streams for downstream tasks when standard cameras fail can be beneficial. To assess the effectiveness of each method in an extrinsic manner, we leverage downstream computer vision tasks including object detection, image classification, and camera calibration.

\noindent \textbf{Object detection.} Object detection is a crucial area of research in computer vision, with numerous applications ranging from autonomous navigation to medical imaging. However, traditional frame-based cameras often fail to capture satisfactory images under low-light conditions, affecting the performance of object detection methods. Since event cameras have a high dynamic range compared to traditional cameras, we evaluated the performance of object detection on low-light images captured by event cameras. For this purpose, we used the MVSEC-NIGHTL21 car detection dataset \cite{hu2021v2e}, derived from the $outdoor\_night\textit{1}\_data$ sequence of MVSEC, captured under night driving conditions. The dataset contains 2,000 labeled intensity images, with 1,600 frames for training and 400 frames for validation. We reconstructed images from the provided event sequence for each method and extracted frames corresponding to those in the MVSEC-NIGHTL21 dataset. We then used YOLOv7 \cite{wang2023yolov7} object detector to detect cars in the reconstructed images and intensity images of the frame camera. We used a model trained on the MS-COCO dataset \cite{lin2014microsoft} for car detection in the images and evaluated the results using the PASCAL VOC metric \cite{everingham2009pascal}, providing the AP score for each method on the dataset. See the supplementary material for sample detection visualizations.

\noindent \textbf{Image classification.} We evaluated the performance of our image reconstruction methods using two image classification datasets: Neuromorphic-Caltech101 (N-Caltech101)\cite{orchard2015converting} and Caltech101\cite{fei2004learning}. N-Caltech101 is a spiking version of the original Caltech101 dataset, containing 100 object classes plus a background class (excluding the ``Faces" class). We trained a ResNet50~\cite{he2016deep} classification model on Caltech101, excluding the ``Faces" class to ensure consistency between the datasets. For each method, we reconstructed images on event streams from N-Caltech101, and ran the ResNet50 model on the reconstructions to evaluate their accuracy on the dataset.

\noindent \textbf{Camera calibration.} It is a critical component of computer vision systems, but traditional calibration techniques for standard frame-based cameras cannot be applied to event cameras due to their asynchronous pixel output. Recently, Muglikar \etal~\cite{muglikar2021calibrate} demonstrated that image reconstruction can be used to apply conventional calibration techniques for accurate event-camera calibration. In this study, we compare the performance of various image reconstruction methods for camera calibration using the \textit{calibration} sequence from the ECD dataset. This sequence consists of an event camera moving in front of a calibration target, and the intrinsic calibration parameters of the DAVIS240C, provided by ECD, serve as the ground truth. We reconstruct image sequences using each method and accordingly obtain intrinsic calibration parameters using the reconstructed images and the \texttt{kalibr} toolbox~\cite{oth2013rolling}. We then measure the mean absolute percentage error (MAPE) of the intrinsic calibration parameters to determine the most effective method.

\section{Evaluation Results and Discussion}
\label{sec:results}

\begin{table*}[!t]
\small
\centering
\setlength{\tabcolsep}{1.1mm}
\renewcommand{\arraystretch}{1.0}
\resizebox{0.92\linewidth}{!}{
\begin{tabular}{l cccc cccc cccc cccc}
\toprule
    &&
    \multicolumn{3}{c}{ECD~\cite{mueggler2017event}} && 
    \multicolumn{3}{c}{MVSEC~\cite{zhu2018multivehicle}} && 
    \multicolumn{3}{c}{HQF~\cite{stoffregen2020reducing}} && 
    \multicolumn{3}{c}{BS-ERGB~\cite{tulyakov2022time}} \\
    \cmidrule{3-5} \cmidrule{7-9} \cmidrule{11-13} \cmidrule{15-17}
    && MSE$\downarrow$ & SSIM$\uparrow$ & LPIPS$\downarrow$ 
    && MSE$\downarrow$ & SSIM$\uparrow$ & LPIPS$\downarrow$ 
    && MSE$\downarrow$ & SSIM$\uparrow$ & LPIPS$\downarrow$ 
    && MSE$\downarrow$ & SSIM$\uparrow$ & LPIPS$\downarrow$ \\
    \midrule
    E2VID~\cite{rebecq2019high} && 
    0.179 & 0.450 & 0.322 && 0.225 & 0.241 & 0.645 && 0.099 & 0.463 & 0.388 && 0.139 & 0.324 & 0.569 \\
    FireNet~\cite{Scheerlinck20wacv} && 
    0.133 & 0.459 & 0.321 && 0.294 & 0.198 & 0.702 && 0.100 & 0.422 & 0.463 && 0.097 & 0.330 & 0.535 \\
    E2VID+~\cite{stoffregen2020reducing} && 
    0.070 & \underline{0.503} & \underline{0.236} && 0.132 & 0.262 & \underline{0.514} && \underline{0.036} & \textbf{0.536} & \textbf{0.255} && \underline{0.076} & \textbf{0.374} & \textbf{0.433} \\
    FireNet+~\cite{stoffregen2020reducing} && 
    \underline{0.062} & 0.452 & 0.289 && 0.219 & 0.212 & 0.570 && 0.045 & 0.472 & 0.323 && 0.091 & 0.318 & 0.482\\
    SPADE-E2VID~\cite{cadena2021spade} && 
    0.091 & 0.461 & 0.337 && 0.138 & \underline{0.266} & 0.591 && 0.079 & 0.405 & 0.514 && 0.091 & 0.339 & 0.643 \\
    SSL-E2VID~\cite{paredes2021back} && 
    0.092 & 0.415 & 0.380 && \underline{0.124} & 0.264 & 0.694 && 0.090 & 0.407 & 0.496 && 0.088 & 0.349 & 0.628 \\
    ET-Net~\cite{weng2021event} && 
    \textbf{0.047} & \textbf{0.552} & \textbf{0.224} && \textbf{0.107} & \textbf{0.288} & \textbf{0.489} && \textbf{0.034} & \underline{0.534} & \underline{0.268} && \textbf{0.072} & \underline{0.366} & \underline{0.445} \\
    \bottomrule
\end{tabular}}

\vspace{-0.3cm}
\caption{\textbf{Full-reference quantitative results on the ECD, MVSEC, HQF, and BS-ERGB datasets}. Here we use between-frames event grouping. The best and second best scores are given in \textbf{bold} and \underline{underlined}.}
\label{tab:quan_res_std}
\end{table*}

\begin{figure*}[t]
	\newcommand{\widthplot}{0.114\textwidth}
	\centering
	\setlength{\tabcolsep}{0.42ex} %
\begin{tabular}{ccccccccc}
    \rotatebox[origin=l]{90}{$\;\;$shapes} &
	\includegraphics[width=\widthplot]{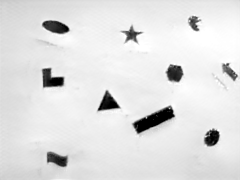} &
	\includegraphics[width=\widthplot]{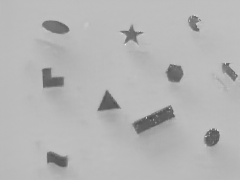} &
	\includegraphics[width=\widthplot]{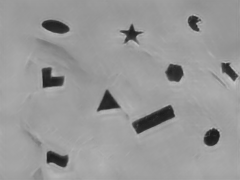} &
	\includegraphics[width=\widthplot]{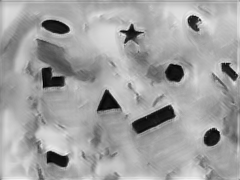} &
	\includegraphics[width=\widthplot]{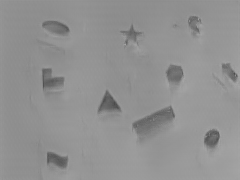} &
	\includegraphics[width=\widthplot]{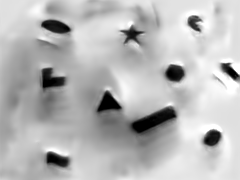} &
	\includegraphics[width=\widthplot]{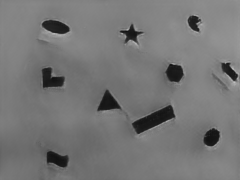} &
	\includegraphics[width=\widthplot]{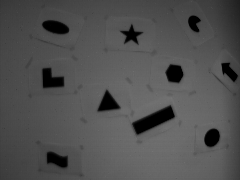} \\	
	
	\rotatebox[origin=l]{90}{out\_day1} &
	\includegraphics[width=\widthplot]{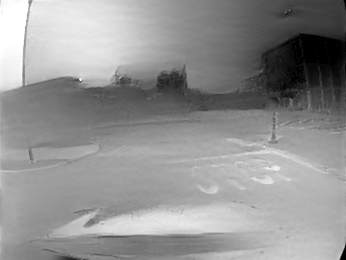} &
	\includegraphics[width=\widthplot]{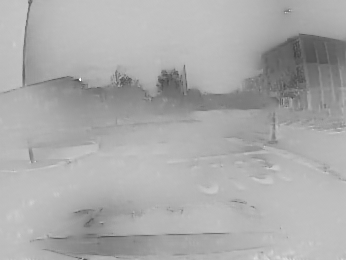} &
	\includegraphics[width=\widthplot]{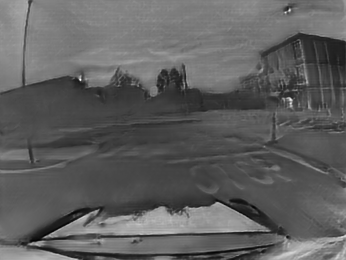} &
	\includegraphics[width=\widthplot]{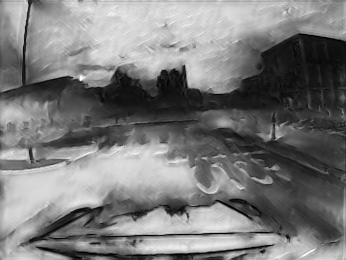} &
	\includegraphics[width=\widthplot]{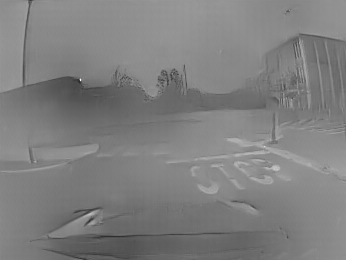} &
	\includegraphics[width=\widthplot]{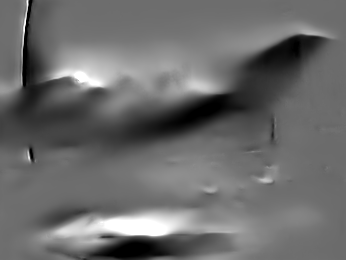} &
	\includegraphics[width=\widthplot]{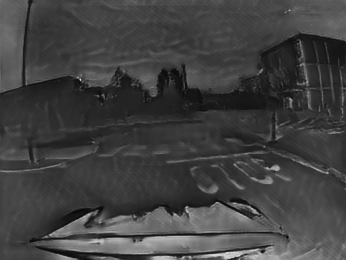} &
	\includegraphics[width=\widthplot]{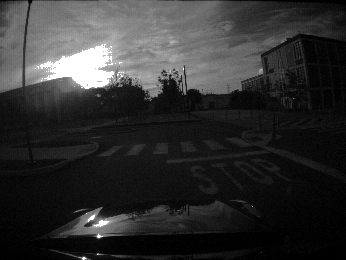} \\

	\rotatebox[origin=l]{90}{$\;\;\;$desk} &
	\includegraphics[width=\widthplot]{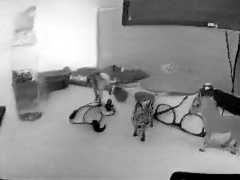} &
	\includegraphics[width=\widthplot]{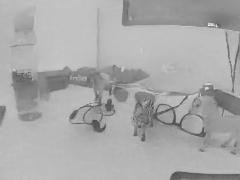} &
	\includegraphics[width=\widthplot]{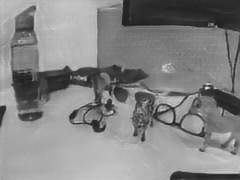} &
	\includegraphics[width=\widthplot]{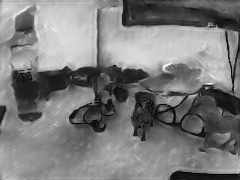} &
	\includegraphics[width=\widthplot]{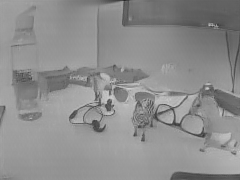} &
	\includegraphics[width=\widthplot]{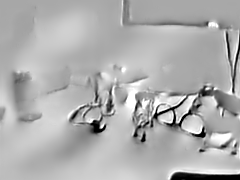} &
	\includegraphics[width=\widthplot]{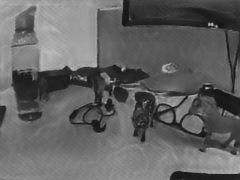} &
	\includegraphics[width=\widthplot]{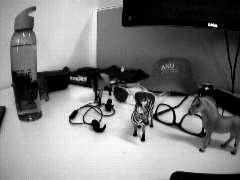} \\

	\rotatebox[origin=l]{90}{$\;\;$street} &
	\includegraphics[width=\widthplot]{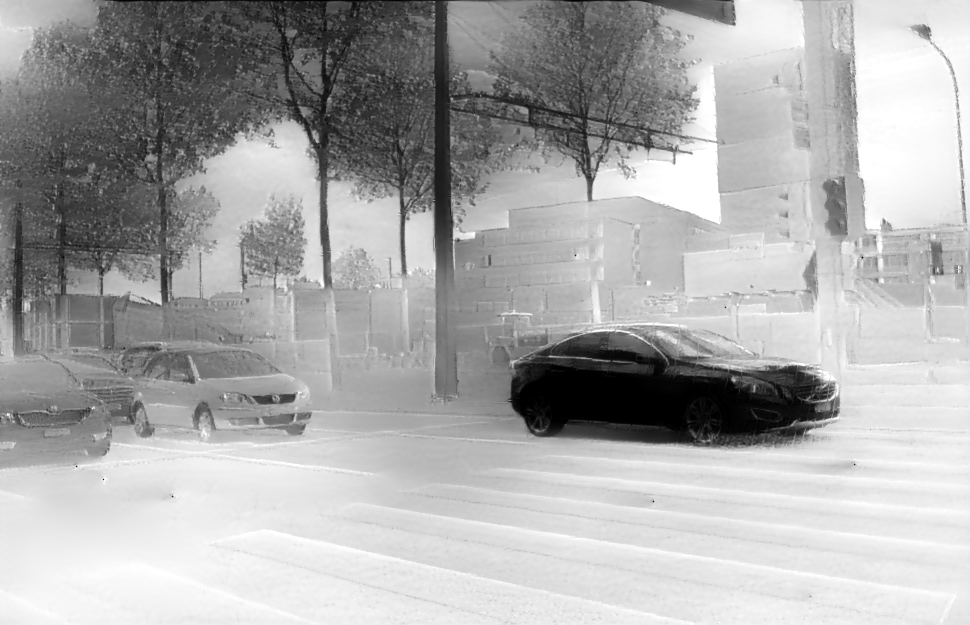} &
	\includegraphics[width=\widthplot]{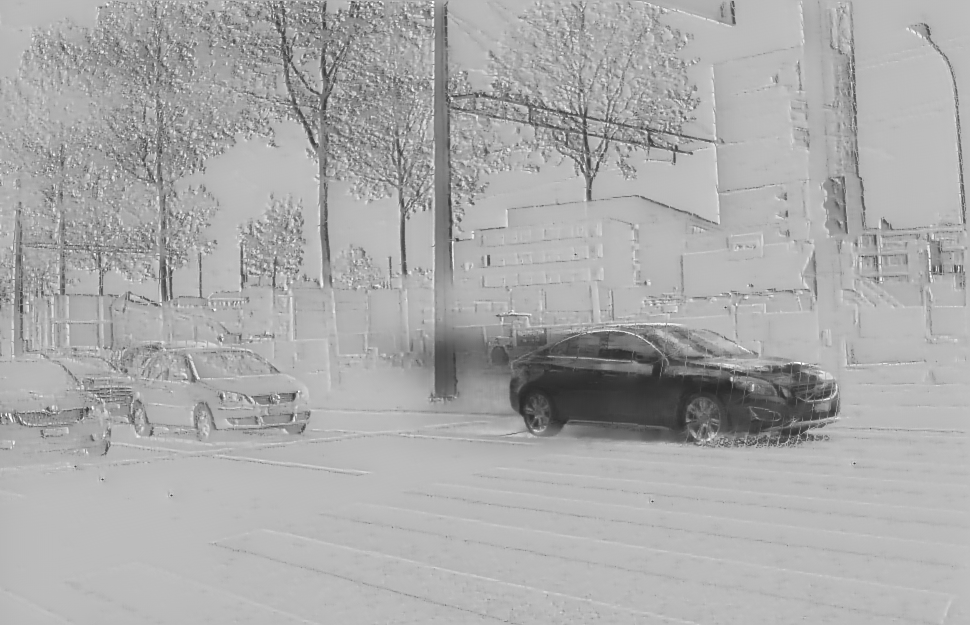} &
	\includegraphics[width=\widthplot]{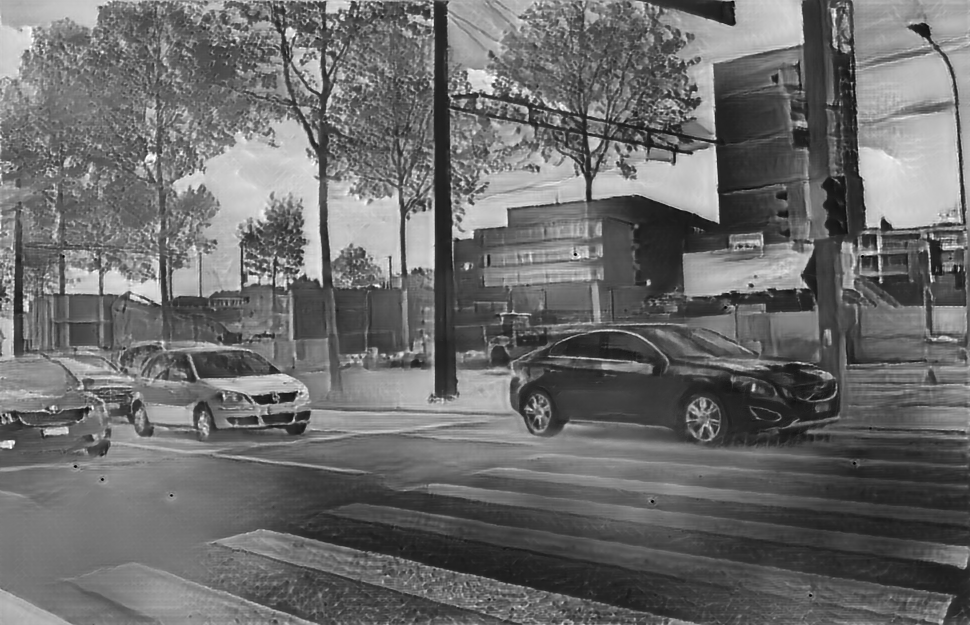} &
	\includegraphics[width=\widthplot]{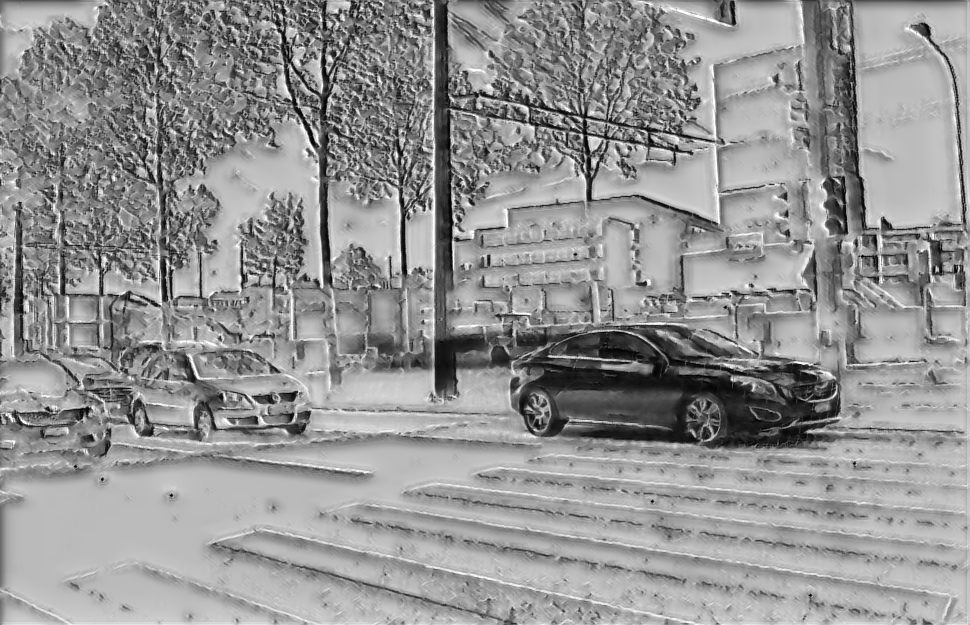} &
	\includegraphics[width=\widthplot]{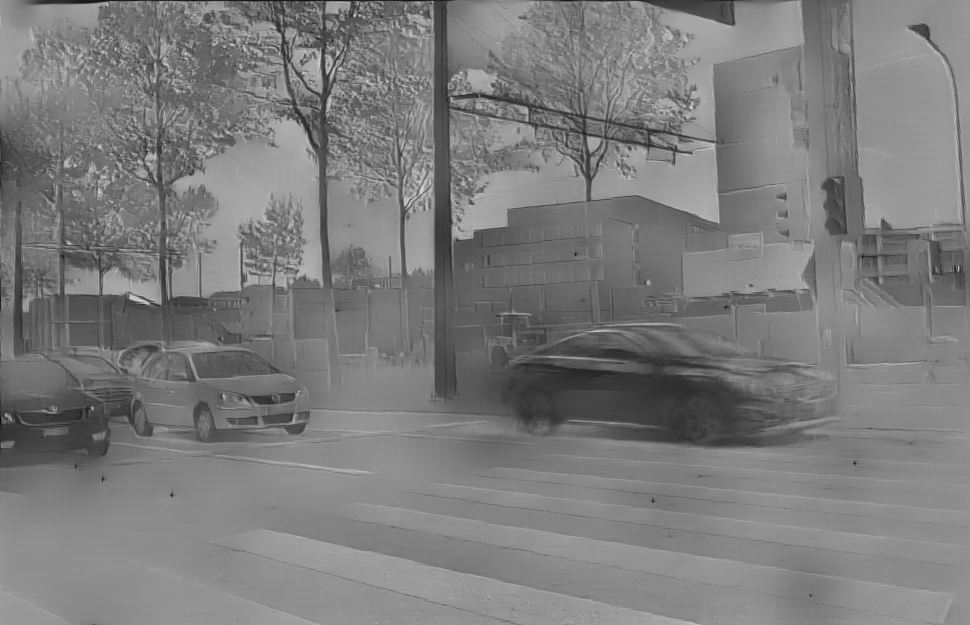} &
	\includegraphics[width=\widthplot]{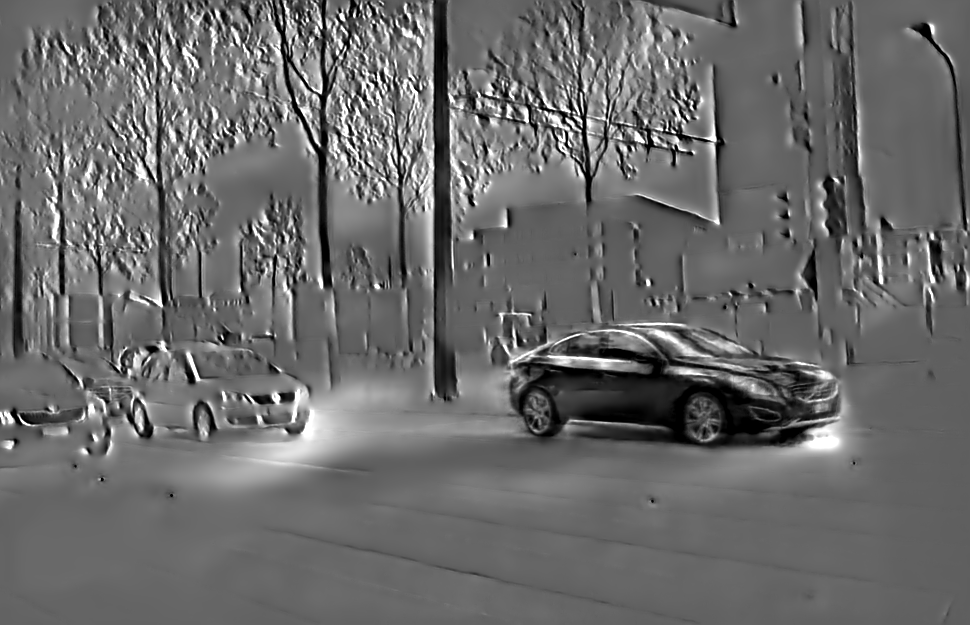} &
	\includegraphics[width=\widthplot]{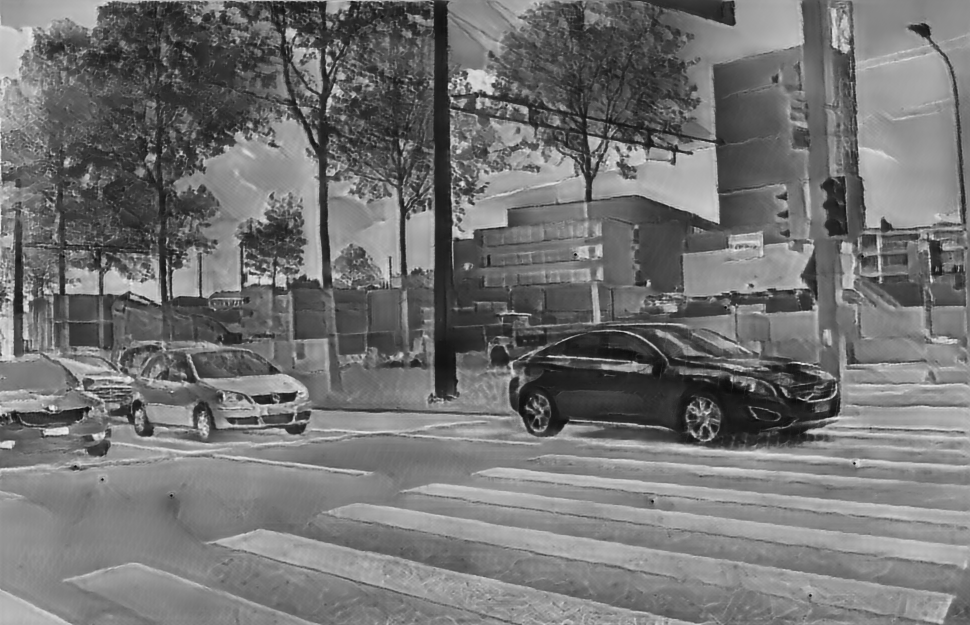} &
	\includegraphics[width=\widthplot]{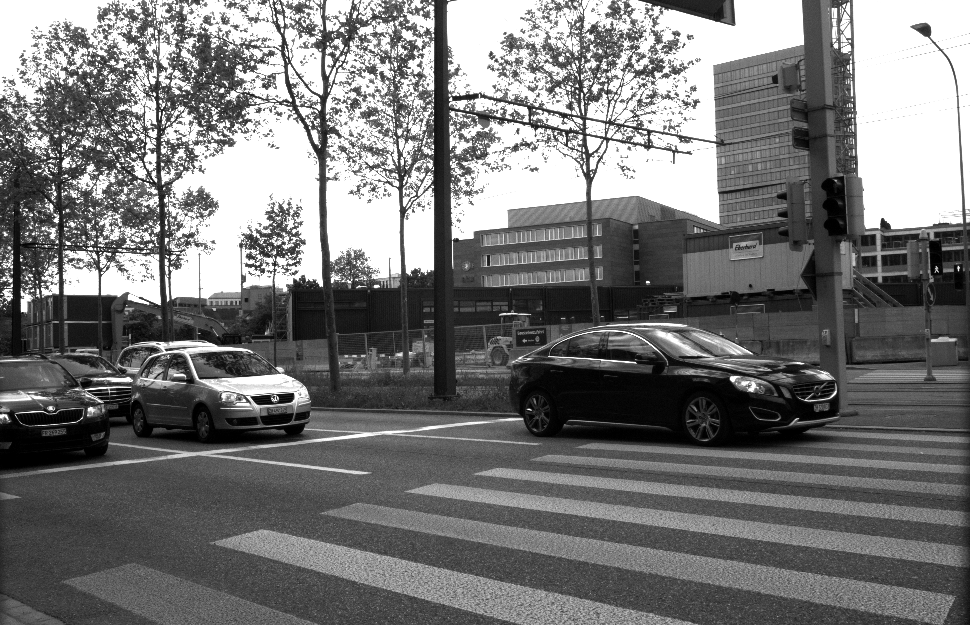} \\

	\rotatebox[origin=l]{90}{$\;\;$selfie} &
	\includegraphics[width=\widthplot]{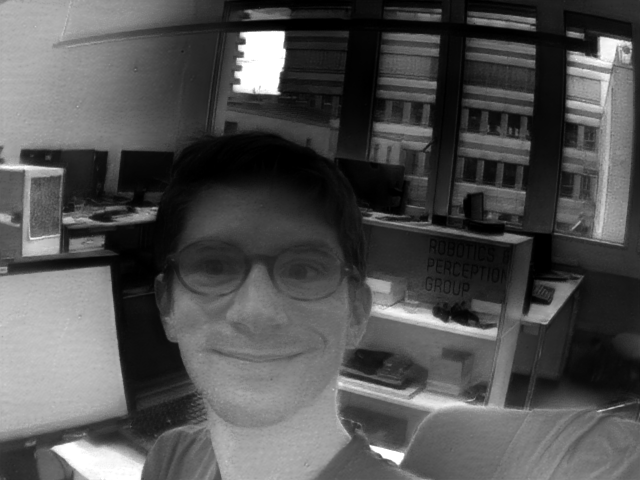} &
	\includegraphics[width=\widthplot]{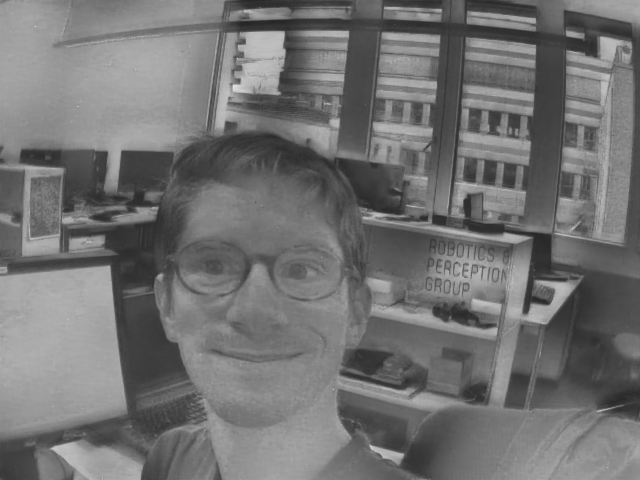} &
	\includegraphics[width=\widthplot]{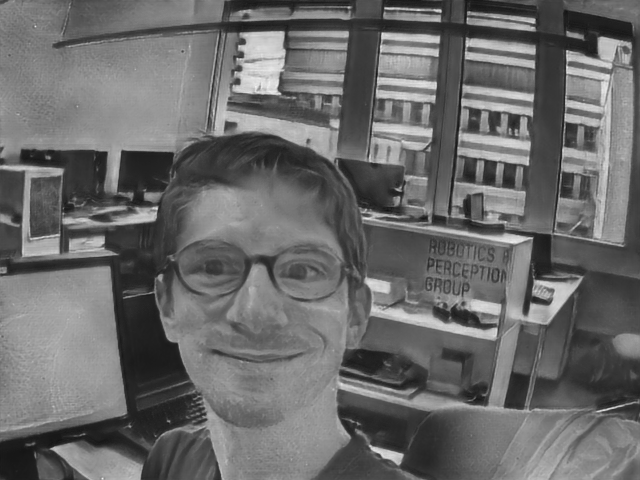} &
	\includegraphics[width=\widthplot]{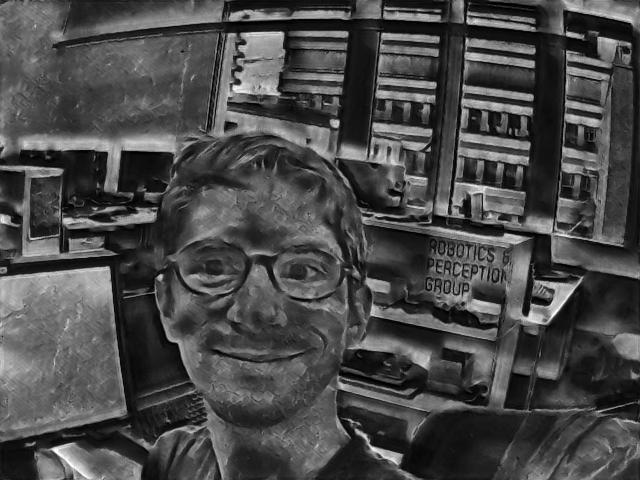} &
	\includegraphics[width=\widthplot]{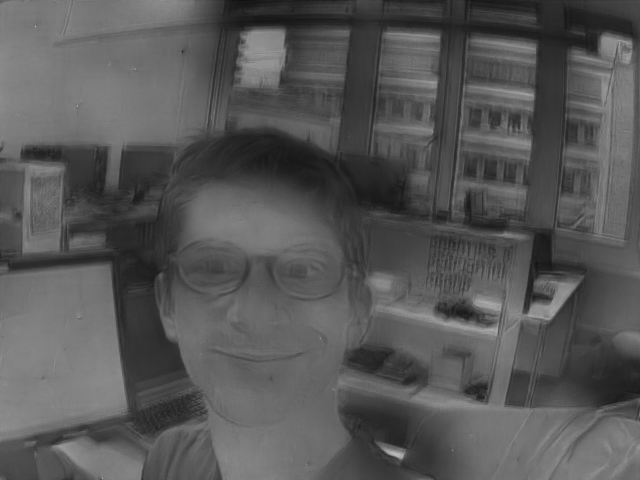} &
	\includegraphics[width=\widthplot]{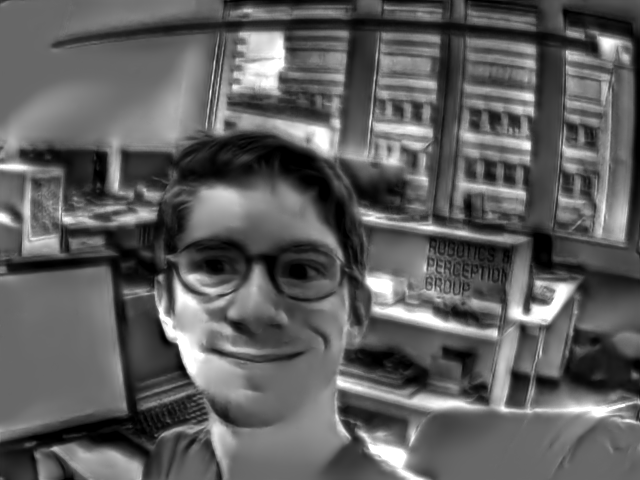} &
	\includegraphics[width=\widthplot]{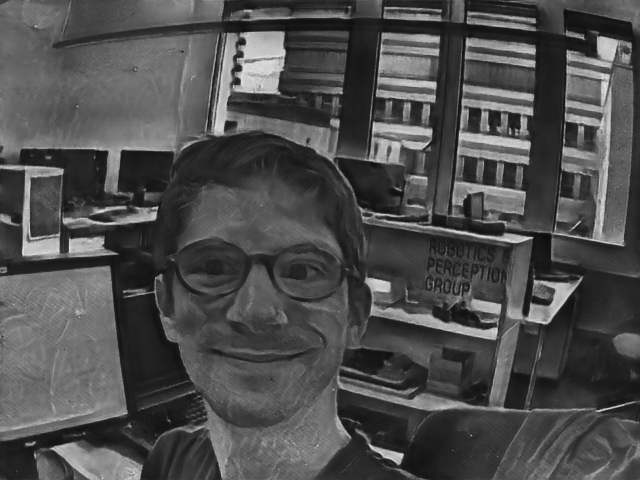} &
	\includegraphics[width=\widthplot]{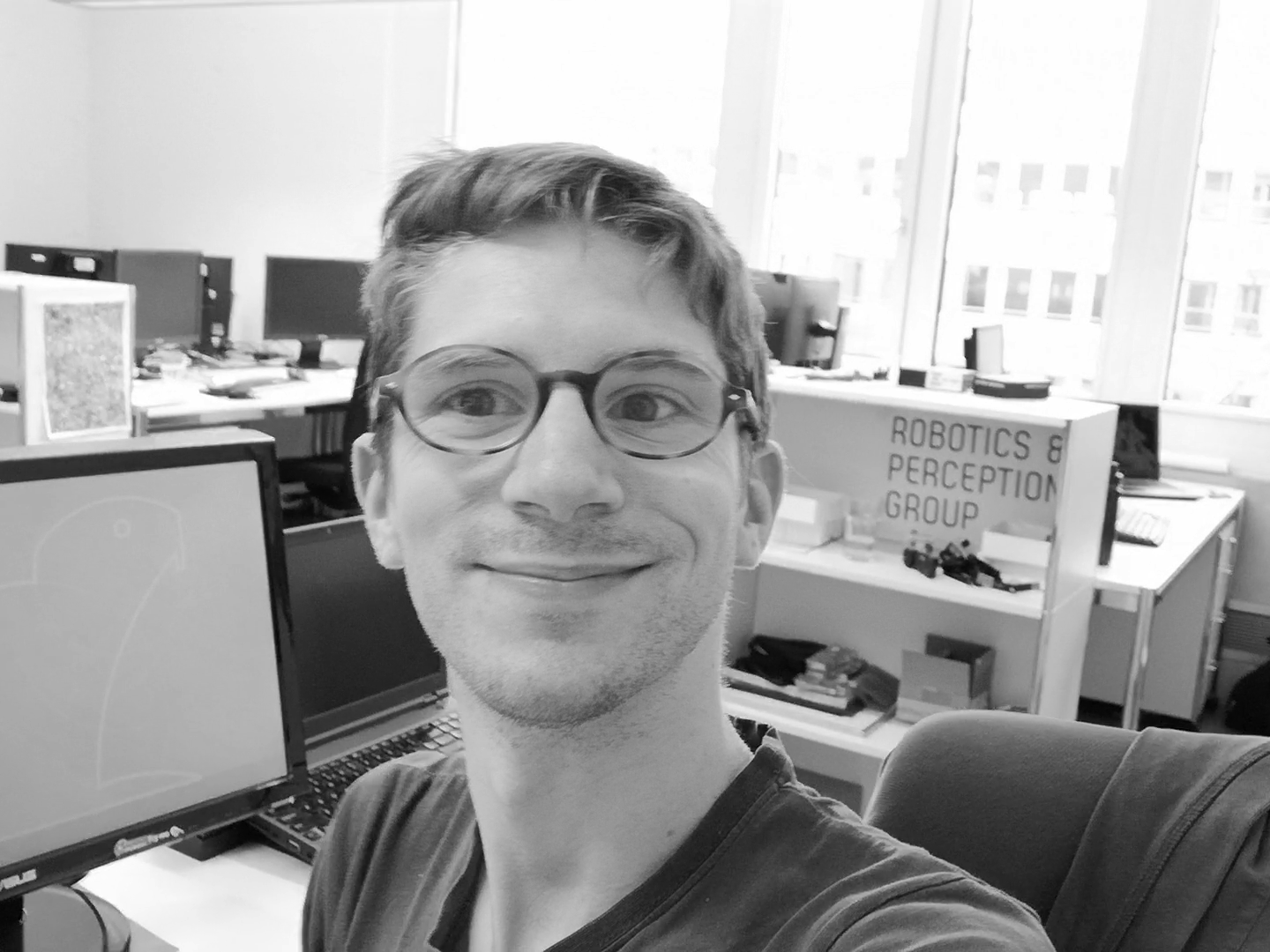} \\
 
	& E2VID & FireNet & E2VID+ & FireNet+ & \makebox[0pt][c]{SPADE-E2VID} & SSL-E2VID & ET-Net & Ground Truth

\end{tabular}
\vspace{-0.3cm}
	\caption{\textbf{Qualitative comparisons}. A sample scene from the ECD, MVSEC, HQF, BS-ERGB, and HDR datasets is given at each row, from top to bottom, respectively (Note that the rightmost image in the last row is a reference frame rather than the groundtruth.)
 \vspace{-0.3cm}}
	\label{fig:qual_eval}
\end{figure*}

\begin{table*}[!t]
\small
\centering
\setlength{\tabcolsep}{1.1mm}
\renewcommand{\arraystretch}{1.0}
\resizebox{0.92\linewidth}{!}{
\begin{tabular}{l cccc cccc cccc }
\toprule
    &&
    \multicolumn{3}{c}{ECD-FAST~\cite{mueggler2017event}} && 
    \multicolumn{3}{c}{MVSEC-NIGHT~\cite{zhu2018multivehicle}} && 
    \multicolumn{3}{c}{HDR~\cite{rebecq2019high}} \\
    \cmidrule{3-5} \cmidrule{7-9} \cmidrule{11-13} 
    && BRISQUE$\downarrow$ & NIQE$\downarrow$ & MANIQA$\uparrow$ 
    && BRISQUE$\downarrow$ & NIQE$\downarrow$ & MANIQA$\uparrow$ 
    && BRISQUE$\downarrow$ & NIQE$\downarrow$ & MANIQA$\uparrow$ \\
    \midrule
    E2VID~\cite{rebecq2019high} && 
    \textbf{14.960} & 7.010 & 0.251 && \textbf{3.155} & 6.401 & 0.319 && \underline{15.719} & 4.347 & 0.310 \\
    FireNet~\cite{Scheerlinck20wacv} && 
    19.951 & 8.094 & 0.271 && 21.338 & 6.073 & 0.292 && 21.539 & 4.085 & 0.308 \\
    E2VID+~\cite{stoffregen2020reducing} && 
    22.626 & \underline{6.735} & 0.263 && 12.300 & \underline{4.312} & 0.311 && 21.340 & 3.903 & 0.305 \\
    FireNet+~\cite{stoffregen2020reducing} && 
    \underline{18.397} & \textbf{5.460} & \textbf{0.304} && \underline{10.020} & \textbf{4.307} & \textbf{0.347} && \textbf{15.680} & \textbf{3.236} & \textbf{0.358} \\
    SPADE-E2VID~\cite{cadena2021spade} && 
    18.917 & 9.992 & \underline{0.288} && 24.012 & 8.486 & 0.314 && 25.835 & 5.567 & 0.286 \\
    SSL-E2VID~\cite{paredes2021back} && 
    46.202 & 9.597 & 0.200 && 49.576 & 10.316 & 0.202 && 55.240 & 6.163 & 0.228 \\
    ET-Net~\cite{weng2021event} && 
    19.699 & 7.529 & 0.283 && 15.533 & 5.228 & \underline{0.329} && 23.526 & \underline{3.643} & \underline{0.335} \\
    \bottomrule
\end{tabular}}
\vspace{-0.2cm}
\caption{\textbf{No-reference quantitative results on challenging sequences involving fast motion, low light, and high-dynamic range}. Here we use between-frames event grouping for ECD-FAST and MVSEC-NIGHT, and fixed-duration event grouping for HDR with a duration of 40 ms. The best and second best results are given in \textbf{bold} and \underline{underlined}.}
\label{tab:quan_res_noref}
\vspace{-0.4cm}
\end{table*}

Table~\ref{tab:quan_res_std} presents the quantitative results of image reconstruction methods on four datasets (ECD, MVSEC, HQF, and BS-ERGB) and using three evaluation metrics (MSE, SSIM, and LPIPS), while \cref{fig:qual_eval} displays qualitative results from sample scenes. The table highlights that the methods ET-Net and E2VID+ are the top performers across all datasets, with ET-Net being overall the most accurate. E2VID+ performs the best on the BS-ERGB dataset, obtaining the lowest LPIPS score and highest SSIM. While the other methods achieve lower performance than these two, some of them still obtain relatively good results on specific datasets. These results demonstrate that the choice of the method can depend on the dataset, highlighting the importance of evaluating methods on multiple datasets to assess their generalization ability.

Table~\ref{tab:quan_res_noref} presents the results of the quantitative analysis on challenging scenarios involving fast motion, low light, and high-dynamic range, assessed by using no-reference metrics BRISQUE, NIQE, and MANIQA. Among the methods compared in the table, FireNet+ achieves the best results. SSL-E2VID obtains the lowest scores in all three metrics compared to other methods. Interestingly, ET-Net, the model that achieves the best scores on standard benchmark datasets in terms of full-reference metrics (\textit{cf.}~Table~\ref{tab:quan_res_std}), performs poorly in these challenging situations. These results suggest that to assess the overall effectiveness of the image reconstruction methods from events, standard benchmark sequences are not enough and further analysis is needed.

\begin{figure*}[t]
    \begin{subfigure}[b]{0.10\textwidth}
        \includegraphics[width=\textwidth]{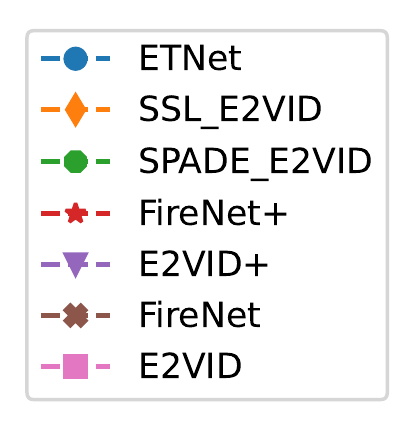} 
        \begin{minipage}{.1cm}
        \vspace{2cm}
        \end{minipage}
    \end{subfigure}
    \begin{subfigure}[b]{0.22\textwidth}
        \centering
        \includegraphics[width=\textwidth]{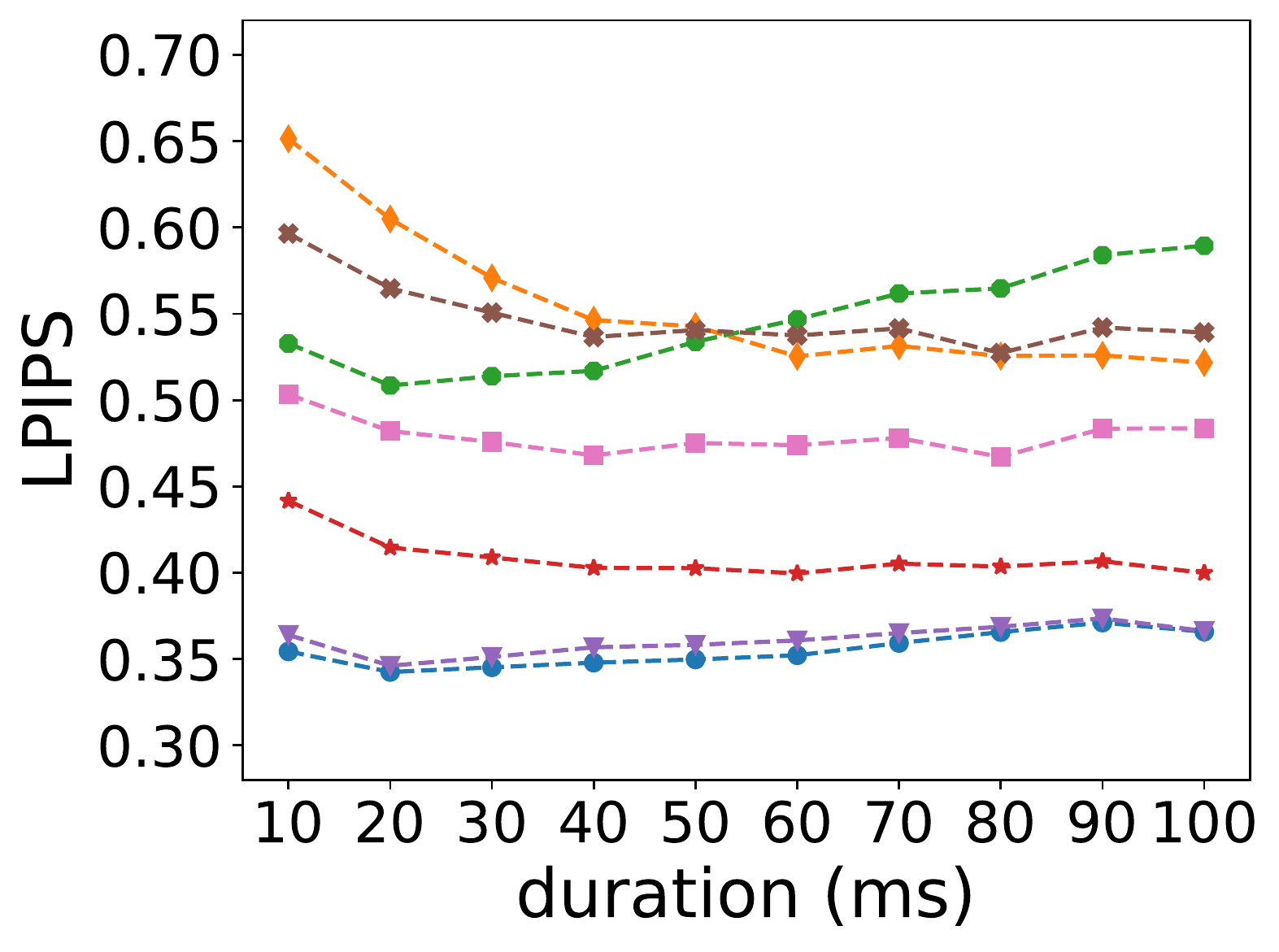} 
        \caption{} \label{fig:robustness_sub1}
    \end{subfigure}
    \begin{subfigure}[b]{0.22\textwidth}
        \centering
        \includegraphics[width=\textwidth]{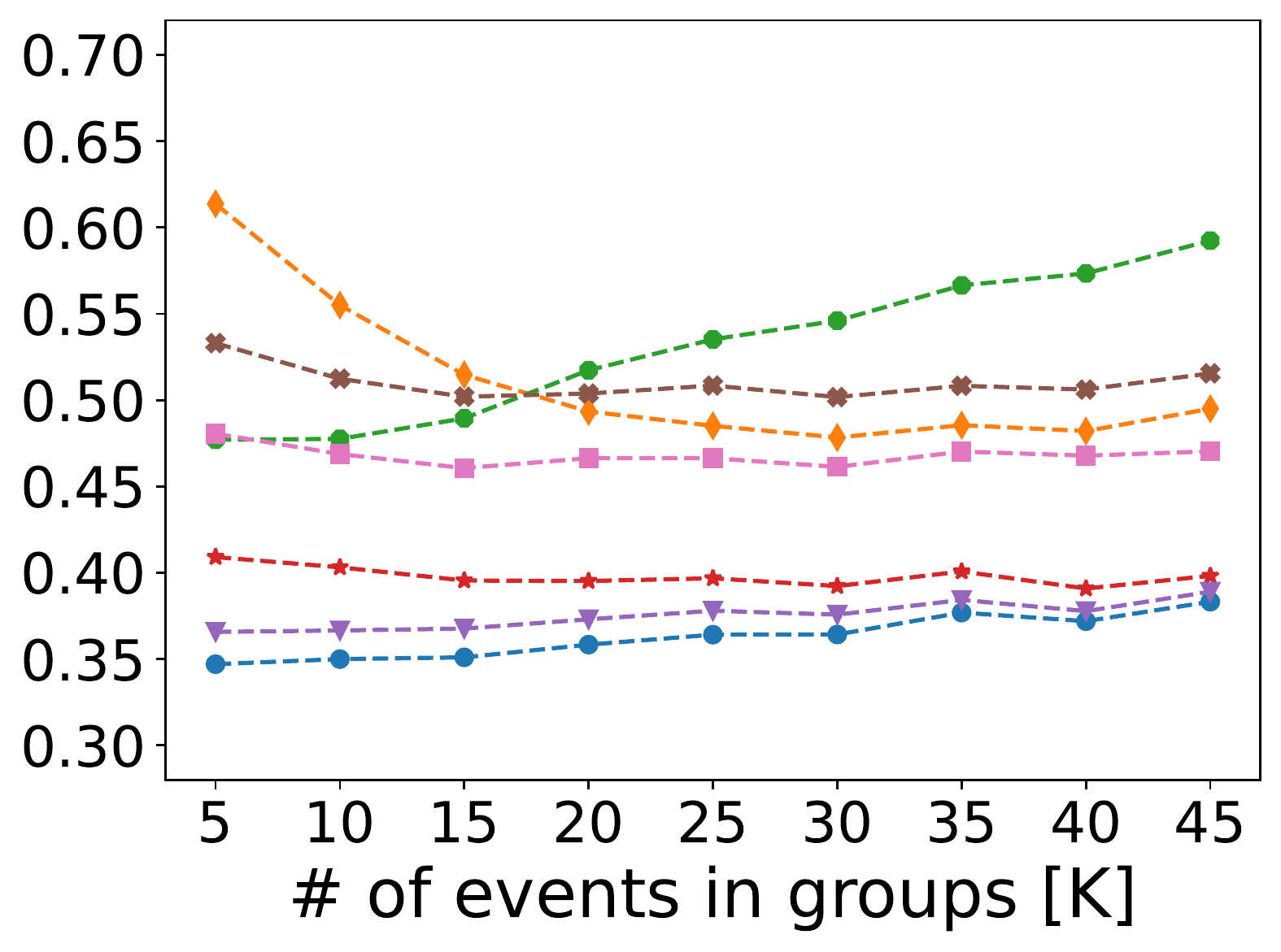} 
        \caption{} \label{fig:robustness_sub2}
    \end{subfigure}
    \begin{subfigure}[b]{0.22\textwidth}
    \centering
        \includegraphics[width=0.97\textwidth]{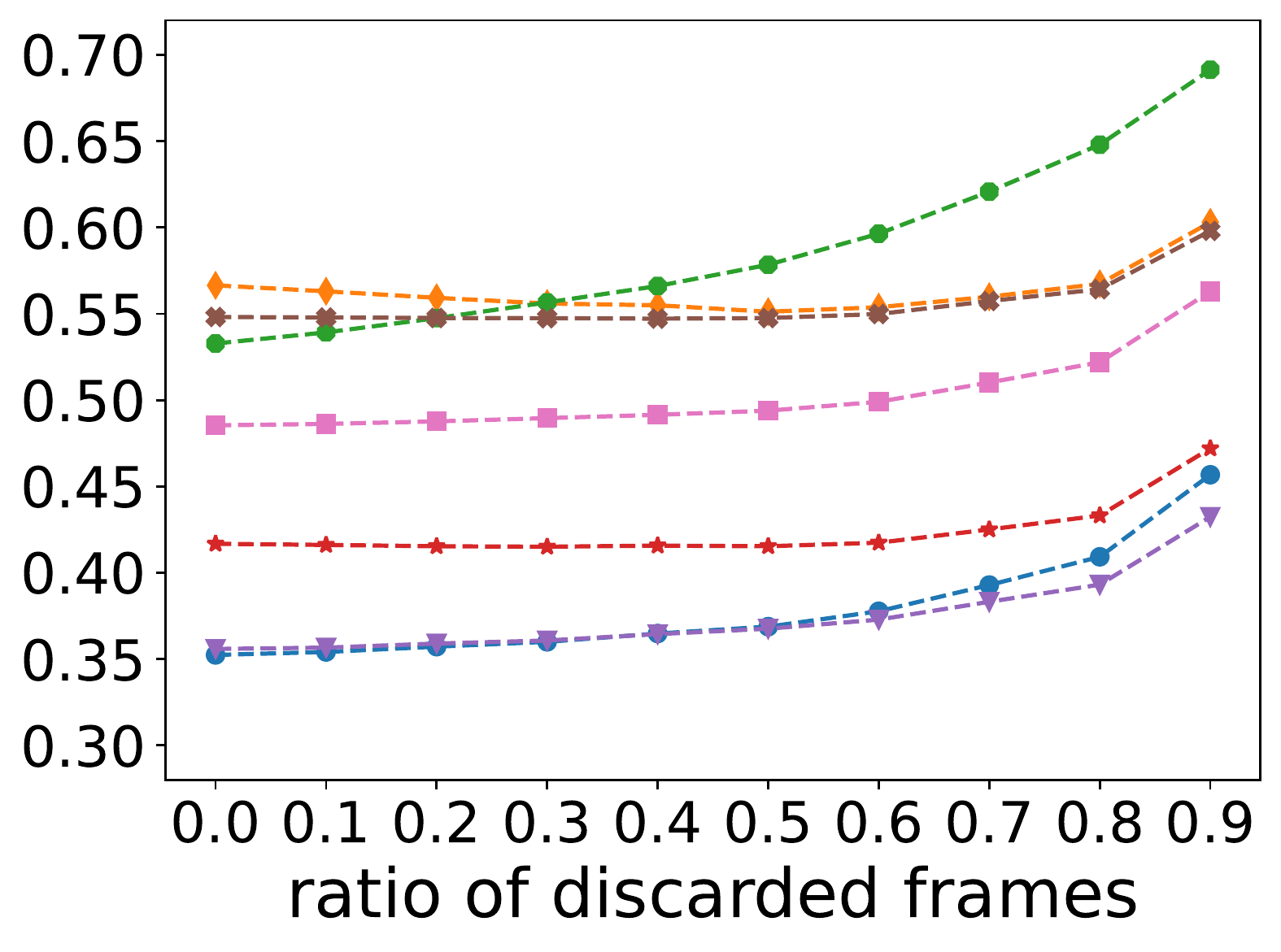} 
        \caption{} \label{fig:robustness_sub3}
    \end{subfigure}
    \begin{subfigure}[b]{0.22\textwidth}
    \centering
        \includegraphics[width=\textwidth]{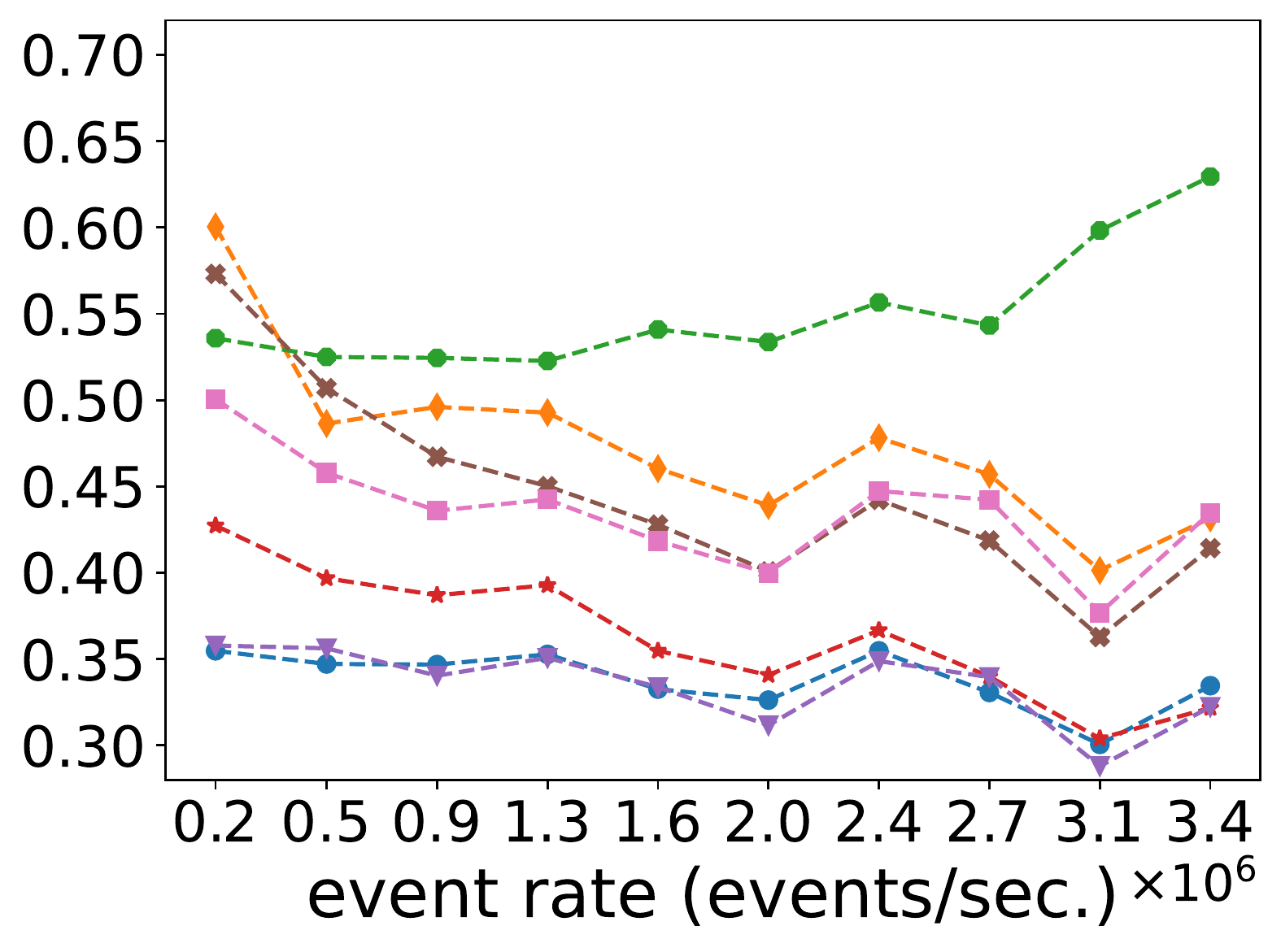} 
        \caption{} \label{fig:robustness_sub4}
    \end{subfigure}
    \vspace{-1.25cm}
    \caption{\textbf{Robustness analysis.} We investigate how factors including (a) image reconstruction rate, (b) event tensor sparsity, (c) temporal irregularity, and (d) event rate affect the performance of each method.} \label{fig:robustness_fig}
    \vspace{-0.3cm}
\end{figure*}

Table~\ref{tab:ds_res} shows the quantitative results of image reconstruction methods on three downstream tasks, including results using ground truth intensity frames as a baseline for comparison. The evaluation metrics employed are AP (Average Precision) for object detection, accuracy for image classification, and MAPE (Mean Absolute Percentage Error) for camera calibration. E2VID achieves the highest score on image classification and second highest score on object detection, with FireNet being the third and second best for these tasks, respectively. E2VID+ obtains a lower score on object detection than these two methods, but still performs well on image classification, achieving the second best score. Its performance on camera calibration is, however, significantly worse than the first two methods. Conversely, SPADE-E2VID has substantially lower performance on image classification, while performing decent on camera calibration. Even though ET-Net achieves state-of-the-art results in full-reference image quality metrics, its downstream task performance is relatively low compared to other methods.

The object detector achieves the highest score when run on the original intensity images. Interestingly, on the camera calibration task, using intensity sequence does not give the minimum MAPE scores. For the image classification task, since the N-Caltech101 dataset does not include intensity images, we leave the accuracy field blank in the last row. In conclusion, choosing a method depends on the specific downstream task: FireNet is superior for night-time vehicle detection, E2VID is the best method for image classification, and FireNet+ is the best performer for camera calibration. However, it is important to consider other factors such as model complexity, training time, and dataset size when choosing a method. Intensity images provide a strong baseline for the object detection task, and further research is needed to improve object detection performance.
\begin{table}[t]
\centering
\setlength{\tabcolsep}{2mm}
\renewcommand{\arraystretch}{1.1}
\resizebox{\linewidth}{!}{
\begin{tabular}{lccccc}
\toprule
 & {Obj. Det.} && {Img. Class.} && {Cam. Cal.} \\ \cmidrule{2-2} \cmidrule{4-4} \cmidrule{6-6}
 {Methods} & {AP (\%)} && {Accuracy (\%)} && {MAPE (\%)} \\ 
\midrule
E2VID~\cite{rebecq2019high} & \underline{53.67} && \textbf{75.99} && 3.26 \\
FireNet~\cite{Scheerlinck20wacv} & \textbf{64.11} && 67.93 && \underline{2.57} \\
E2VID+~\cite{stoffregen2020reducing} & 52.15 && \underline{70.47} && 6.26 \\
FireNet+~\cite{stoffregen2020reducing} & 28.83 && 47.17 && \textbf{1.70} \\
SPADE-E2VID~\cite{cadena2021spade} & 35.80 && 19.53 && 2.89 \\
SSL-E2VID~\cite{paredes2021back} & 52.03 && 60.68 && 8.58 \\
ET-Net~\cite{weng2021event} & 47.87 && 66.52 && 3.88 \\
\midrule
Ground Truth Frames & 72.36 && -- && 4.53 \\
\bottomrule
\end{tabular}}
\vspace{-0.25cm}
\caption{\textbf{Quantitative results on downstream tasks.} The best and second best results are highlighted in \textbf{bold} and \underline{underlined}.}
\vspace{-0.6cm}
\label{tab:ds_res}
\end{table}

\cref{fig:robustness_fig} shows plots of mean LPIPS scores for robustness analysis. As the event grouping duration increases~(\ref{fig:robustness_sub1}), some of the worse-performing methods start to improve while the best-performing methods maintain their performance. When the number of events in groups increases~(\ref{fig:robustness_sub2}), the performance of SPADE-E2VID decreases significantly, while a decrease in the number of events reduces the performance of SSL-E2VID and E2VID. The other methods remain fairly robust to changes in this setting. Interestingly, as we discard 10\% of the ground truth frames~(\ref{fig:robustness_sub3}), all the performances improve, which may be an indication of a sub-optimal event grouping in the original setting. As the discard ratio is increased above 0.1, the performances decrease significantly, except for E2VID. SPADE-E2VID is susceptible to event rate increase~(\ref{fig:robustness_sub4}), while this change is beneficial for some other methods.

We also analyzed the computational complexity of each method, which is presented in the supplementary material.

\section{Conclusion}
\label{sec:conclusion}
This paper presents a framework called EVREAL, which provides a unified evaluation scheme for event-based video reconstruction methods. EVREAL can serve as a valuable resource for researchers and practitioners working in event-based vision. %
In this study, we utilized EVREAL to analyze state-of-the-art models and yielded insightful observations on their performance under varying settings, challenging scenarios, and downstream tasks.
These models, however, require certain event representations as their inputs, making evaluating them with different event representations impractical. This could be considered a limitation of the current work. Future work will include incorporating a temporal consistency metric, expanding our test datasets, exploring additional downstream tasks, and developing color image reconstruction capabilities in conjunction with model training. Overall, we believe that our work will contribute to the development of more effective and robust event-based video reconstruction models. %

\section{Acknowledgments}
This work was supported in part by KUIS AI Research Award, TUBITAK-1001 Program Award No. 121E454, and BAGEP 2021 Award of the Science Academy to A. Erdem.

\clearpage
\newpage
{\small
\bibliographystyle{ieee_fullname}
\bibliography{ms}
}

\end{document}

\title{EVREAL: Towards a Comprehensive Benchmark and Analysis Suite for Event-based Video Reconstruction \\~\\ Supplementary Material}

\author{Burak Ercan \textsuperscript{1,2}
\quad\quad Onur Eker \textsuperscript{1,2}
\quad\quad Aykut Erdem \textsuperscript{3,4}
\quad\quad Erkut Erdem \textsuperscript{1,4} \vspace{0.2cm} \\
\textsuperscript{1} Hacettepe University, Computer Engineering Department ~~ \textsuperscript{2} HAVELSAN Inc.  \\
\textsuperscript{3} Ko\c{c} University, Computer Engineering Department ~~ \textsuperscript{4} Ko\c{c} University, KUIS AI Center \\
}
\maketitle

In this supplementary document, we provide additional
material to complement the main paper. First, we present recent related work on event-based video reconstruction, especially focusing on their evaluation details (\cref{sec:related}). Second, we share the details of the event representation that we have employed (\cref{sec:evrep}). Third, we share implementation details of the evaluation metrics we considered in our analysis (\cref{sec:metrics}). Next, we provide the overview of the datasets being used in our proposed EVREAL framework (\cref{sec:datasets}). Then, we present the details and results from the computational complexity analysis that we performed (\cref{sec:computational}). Finally, we share additional qualitative results from several datasets (\cref{sec:add_qual}).

\section{Related Work}
\label{sec:related}

 In recent years, there has been a surge of methods aiming to reconstruct intensity images from events, each taking into account different assumptions and employing distinct processing methodologies. Early approaches were limited, often relying on basic assumptions such as known or restricted camera movement, static scenes, or brightness constancy. More recent methods utilize deep neural networks and incorporate natural image priors in their models to achieve better results. Here, we limit our discussion to these recent methods and especially focus on their evaluation details.

Wang \etal~\cite{wang2019event_cvpr} proposed a conditional GAN based model, in which input events are represented by means of spatio-temporal voxel grids. Their evaluation setup includes a small amount of data containing 1000 intensity frames taken from both real and simulated datasets, including the sequences from \cite{bardow2016simultaneous}. 
They compared their method against~\cite{bardow2016simultaneous, munda2018real} for sequences without any ground truth outputs, by utilizing the no-reference metric BRISQUE~\cite{mittal2012no}.
The authors do not share their evaluation code.

Rebecq \etal~\cite{rebecq2019events, rebecq2019high}  introduced a recurrent fully convolutional network, named E2VID.
The authors used a selection of seven sequences from the ECD~\cite{mueggler2017event} dataset, using a fixed number of events to form event voxel grids and a tolerance of 1 ms to match the reconstructions with ground truth frames. To improve the output quality, they applied robust normalization as a post-processing step and then performed local histogram equalization before computing scores for MSE, SSIM and LPIPS~\cite{zhang2018unreasonable}. They compared their approach against \cite{bardow2016simultaneous} and \cite{munda2018real}. They also reported a temporal consistency score that requires a ground truth optical flow map between each frame. To obtain this, they used an off-the-shelf frame-based optical flow network~\cite{ilg2017flownet}, which has its own prediction errors. The researchers conducted experiments on challenging scenarios involving rapid motion, low-light conditions and high dynamic range, without providing any quantitative scores. Additionally, they reported color image reconstruction results from the event data available in CED dataset~\cite{scheerlinck2019ced}, without providing any quantitative analysis. 

Rebecq \etal also evaluated their method on four downstream tasks, including image classification, visual-inertial odometry, object detection, and monocular depth estimation~\cite{rebecq2019events, rebecq2019high}. To perform these tasks, they fed reconstructed frames as inputs to task-specific frame-based methods and reported either qualitative or quantitative results. For instance, for object classification, they used events from N-MNIST~\cite{orchard2015converting}, N-Caltech101~\cite{orchard2015converting}, and N-Cars~\cite{sironi2018hats} datasets, and provided accuracy scores achieved by a ResNet-18~\cite{he2016deep} network. Similarly, for visual-inertial odometry, they employed events from the ECD dataset, and investigated mean translation errors obtained via VINS-Mono~\cite{qin2018vins}. For object detection and monocular depth estimation, they used YOLOv3~\cite{redmon2018yolov3} and MegaDepth~\cite{li2018megadepth}, respectively, and only shared qualitative results in a supplementary video. Additionally, they analyzed the computational efficiency of their approach by reporting the frame synthesis time. The authors do not release their evaluation code publicly.

Scheerlinck \etal~\cite{Scheerlinck20wacv} proposed FireNet, a lightweight recurrent network, as a replacement for E2VID, and demonstrated that it can attain similar performance with much less memory consumption and faster inference. In their evaluation setup, they followed the methodology in \cite{rebecq2019high}, and performed experiments on the selected frames from the sequences in the ECD dataset. They utilized a fixed number of events to form event voxel grids, and applied local histogram equalization to reconstructions and ground truth frames before estimating quantitative metrics such as MSE, SSIM, and LPIPS. Additionally, they performed qualitative analysis on color image reconstruction and challenging scenarios involving high-dynamic range and fast motion. They focused on evaluating computational efficiency and compared several resolutions on GPU and CPU by examining the number of model parameters, memory consumption, FLOPs, and inference times. However, they did not conduct any downstream task experiments, and their evaluation codes are not made publicly available.

Stoffregen \etal~\cite{stoffregen2020reducing} proposed an enhanced version of E2VID, named E2VID+, by retraining it on synthetic training data exhibiting similar statistics with real-world test data. They also employed the same strategy for improving the FireNet architecture, resulting in FireNet+. They evaluated their methods on a larger set of real-world sequences from three datasets, namely ECD and MVSEC~\cite{zhu2018multivehicle} datasets, and their proposed HQF dataset. For ECD and MVSEC, they used the sequences commonly used in earlier work, and reported MSE, SSIM, and LPIPS scores. They always had a matching ground truth frame for each reconstruction, as they used events between each consecutive ground truth frame to form voxel grids. It is not clear whether they applied normalization or histogram equalization before calculating these scores. Moreover, they did not perform any experiments on challenging scenarios or downstream tasks, nor did they perform computational efficiency analysis. The evaluation code is not publicly available.

Cadena \etal~\cite{cadena2021spade} proposed SPADE-E2VID, which integrates spatially-adaptive denormalization (SPADE)\cite{park2019semantic} layers into the E2VID architecture to enhance the quality of the reconstructed videos. The authors evaluated their approach using seven sequences from the ECD dataset, starting from the very first frames of each sequence, and reported MSE, SSIM, and LPIPS scores for quantitative comparison with E2VID and FireNet. They also introduced an RMS contrast metric to demonstrate that their method produces higher contrast reconstructions. To assess temporal consistency, they used a different off-the-shelf frame-based optical flow network\cite{ranjan2017optical} and reported the corresponding scores. In addition, they performed object detection analysis on a single sequence of the ECD dataset, using events and YOLOv4~\cite{bochkovskiy2020yolov4} to process reconstructed frames. They estimated ground truth object labels for two object classes by applying the same object detection network to ground truth intensity images and shared average precision scores for this downstream task accordingly. They analyzed the computational efficiency of their approach by reporting reconstruction time for inputs with various resolutions. While they released an evaluation code, we were unable to reproduce their results with it.

Weng \etal~\cite{weng2021event} improved the E2VID architecture by adding a Transformer-based module to better exploit the global context of event tensors, thus naming their model as ET-Net. Their experiments were conducted using the ECD, MVSEC, and HQF datasets, with the same sequence cuts as in~\cite{stoffregen2020reducing}. In their experiments, events between consecutive ground truth frames were used to form voxel grids. To evaluate their approach, they calculated MSE, SSIM, and LPIPS scores, without any normalization or histogram equalization applied to the reconstructed images. They compared their method with E2VID, E2VID+, FireNet, and FireNet+, and shared qualitative results on challenging scenarios involving high-dynamic-range and rapid motion in their supplementary material. However, they did not perform a computational efficiency analysis or an experiment on a downstream task. The authors provided an open-source evaluation code, and we are able to use to reproduce their results.

Paredes-Vallés and de Croon~\cite{paredes2021back} proposed a self-supervised learning method called SSL-E2VID, which employs the event-based photometric constancy assumption \cite{gallego2015event} to estimate optical flow and intensity images simultaneously. As done in earlier work, events between each consecutive ground truth frame were used to form voxel grids. Their experiments were conducted on ECD and HQF datasets, and they made quantitative comparisons with E2VID, E2VID+, FireNet, and FireNet+. Local histogram equalization was employed before calculating quantitative scores. Since they did not introduce a new architecture, computational efficiency analysis was not performed. Qualitative results were given also for challenging scenarios such as high-dynamic-range and high-speed. No downstream task analysis was performed, and their evaluation code was not made publicly available.

Zhu et al.~\cite{zhu2022event} proposed a spiking neural network architecture that achieves comparable performance to E2VID, E2VID+, FireNet, and SPADE-E2VID with higher computational efficiency. They used the ECD, MVSEC, and HQF datasets in their evaluation and reported quantitative scores using MSE, SSIM, and LPIPS metrics. In reconstructing intensity images, they used the events between each consecutive ground truth frame as input. They applied histogram equalization before calculating these scores. In addition, they provided an analysis of energy consumption. However, they did not release an open-source evaluation code.

Zhang \etal~\cite{zhang2022formulating} presented a novel approach for event-based image reconstruction by formulating it as a linear inverse problem based on optical flow. They conducted a quantitative comparison with E2VID, E2VID+, and SSL-E2VID using MSE, SSIM, and LPIPS metrics. They focused on test sequences with limited camera motion, specifically selected from the ECD dataset, and utilized events from N-Caltech101~\cite{orchard2015converting} dataset. They aligned reconstructions with respective reference frames using Enhanced Correlation Coefficient Maximization~\cite{evangelidis2008parametric}. They reported median scores for each sequence instead of mean scores and presented distribution plots of scores of each method on various sequences. They also analyzed the effect of histogram equalization on quantitative scores and emphasized the importance of taking various factors into account while interpreting these scores. They showcased their method's ability to perform color reconstruction and demonstrated temporal consistency on two example frames from the DSEC dataset~\cite{gehrig2021dsec}. They did not conduct experiments on downstream tasks and did not share their evaluation code.

\section{Details of Event Representation} \label{sec:evrep}

Following the common practice in the literature, we form voxel grids from grouped events in order to utilize deep CNN architectures for event-based data. Let $G_k$ be a group of events that span a duration of $\Delta T$ seconds, $T_k$ be the starting timestamp of that duration, and $B$ be the number of temporal bins used to discretize the timestamps of continuous-time events in the group. The voxel grid $V_k \in {\rm I\!R}^{W \times H \times B}$ for that group is formed by normalizing the timestamps of events from the group to the range $[0, B-1]$. Each event contributes its polarity to the two temporally closest voxels using a linearly weighted accumulation similar to bilinear interpolation. Specifically, the voxel grid is computed as follows:
\begin{equation}
\label{eq:event_representation_voxel}
V_k(x,y,t) = \sum_{\substack{i}} {p_i \max(0, 1-|t - t^*_i|) {\delta}(x-x_i,y-y_i)}
\end{equation}
where $\delta$ is the Kronecker delta that selects the pixel location, and $t^*_i$ is the normalized timestamp which is calculated as:
\begin{equation}
\label{eq:event_representation_ts_norm}
t^*_i = (B-1)(t_i - T_k) / (\Delta T)
\end{equation}
In all our experiments, we set the number of temporal bins $B$ to 5.

\section{Implementation Details for Quantitative \mbox{Image} Quality Metrics}
\label{sec:metrics}
\vspace{0.25cm}
\noindent \textbf{MSE.} The Mean Squared Error is a commonly used metric that does not require any parameters. When comparing two images, the only factor that can impact the MSE result is the range of pixel values that the images possess. We calculate the MSE using floating-point pixel values within the range $[0,1]$. Lower MSE values indicate better results.
    
\vspace{0.25cm}
\noindent \textbf{SSIM.} We utilize the \texttt{scikit-image} image processing library's~\cite{van2014scikit} implementation for structural similarity. We adjust the parameters to use the Gaussian weighting scheme explained in~\cite{wang2004image}. Like MSE, we compute SSIM using images with floating point pixel values in the range of $[0,1]$. Higher scores of SSIM indicate better results.

\vspace{0.25cm}
\noindent \textbf{LPIPS.} For LPIPS, we use the implementation in IQA-PyTorch toolbox\cite{pyiqa}\footnote{\label{iqa}The code is accessible from \url{https://github.com/chaofengc/IQA-PyTorch}}, v0.1.10. We employ the default settings, using the LPIPS variant that uses the pre-trained AlexNet~\cite{krizhevsky2017imagenet} network. The implementation supports 3-channel RGB images. Therefore, we convert intensity images into RGB images by concatenating three copies of the grayscale image along the channel axis before calculating the scores. In the LPIPS score calculation, a lower score indicates better quality.

\vspace{0.25cm}
\noindent \textbf{BRISQUE.} 
For BRISQUE~\cite{mittal2012no}, we again use the implementation in IQA-PyTorch toolbox\cite{pyiqa}, v0.1.10, with default settings. Similarly, we concatenate three copies of the grayscale image along the channel axis. Lower BRISQUE scores are better.

\vspace{0.25cm}
\noindent \textbf{NIQE.}
For NIQE~\cite{mittal2012making}, we again use the implementation in \mbox{IQA-PyTorch toolbox\cite{pyiqa}}, v0.1.10, with default settings, and concatenating three copies of the grayscale image along the third dimension. Lower NIQE scores are better. 

\vspace{0.25cm}
\noindent \textbf{MANIQA.}
For MANIQA~\cite{yang2022maniqa}, we follow the same approach and use the implementation in IQA-PyTorch toolbox\cite{pyiqa}, v0.1.10, with default settings. In contrast to the above metrics, higher MANIQA scores imply higher image quality. 

\section{Dataset Details}
\label{sec:datasets}
\vspace{0.25cm}
\noindent \textbf{Event Camera Dataset (ECD).}
Following the practice explained in~\cite{rebecq2019high}, we use seven different sequence with diverse characteristics from the ECD dataset~\cite{mueggler2017event}. These sequences are short, taken indoors, and mostly contain simple scenes of office environments with stable objects. The data was captured by a DAVIS240C sensor \cite{brandli2014240}, which is mostly moving with 6 degrees of freedom (DOF) and with increasing speed. The camera generates events and frames from the same pixel array, which has a spatial resolution of $240 \times 180$ pixels. The ground truth intensity frames are available at an average rate of 22 Hz.

To allow methods to generate meaningful results, we exclude the initial few seconds of each sequence from quantitative evaluation. Additionally, when using full-reference metrics, as commonly done in earlier work, we do not include the latter parts of the sequences as they may contain motion blur due to the increased speed of camera movement. However, when evaluating with no-reference metrics, we specifically concentrate on these sections that have fast camera movement, to which the corresponding ground truth intensity images are of lower quality. %

\vspace{0.25cm}
\noindent \textbf{Multi Vehicle Stereo Event Camera (MVSEC) dataset.}
The MVSEC dataset~\cite{zhu2018multivehicle} contains longer sequences captured by a pair of DAVIS 346B cameras, each having a spatial resolution of $346 \times 260$ pixels. These sequences depict both indoor and outdoor environments. 
To evaluate the quality of the videos generated by the methods using full-reference metrics, we followed the approach taken by \cite{stoffregen2020reducing} and considered six commonly used sequences from this dataset. Four of these sequences were captured indoors by a flying hexacopter, while the remaining two were taken outdoors during the daytime from a driving vehicle. The average rate of ground truth intensity frames was approximately 30 Hz for indoor sequences and 45 Hz for outdoor sequences. 
Additionally, we used three night sequences from this dataset, each captured from a vehicle as well, for our experimental evaluation involving no-reference metrics as the ground truth frames at night-time tend to be underexposed.

\vspace{0.25cm}
\noindent\textbf{High-Quality Frames (HQF) dataset.}
The HQF dataset~\cite{stoffregen2020reducing} contains fourteen sequences that exhibit a wide range of different motion behaviors, including static, slow, and fast camera motion, and cover both indoor and outdoor scenes. Two different DAVIS240C cameras are used to capture the data, providing distinct noise and contrast threshold characteristics. The cameras generate events and intensity frames from the same $240 \times 180$ pixel array. The scenes and camera parameters are adjusted to ensure that the ground truth frames are well-exposed and have minimal motion-blur. The average rate of ground truth intensity frames is 22.5 Hz. We use the entire sequences from this dataset for evaluation using full-reference quantitative metrics.

\vspace{0.25cm}
\noindent\textbf{Beam Splitter Event and RGB (BS-ERGB) Dataset.}
The BS-ERGB Dataset~\cite{tulyakov2022time} is originally collected for the event-based video frame interpolation task. The dataset consists of events recorded by a Prophesee Gen4M event camera~\cite{finateu20205} having a spatial resolution of 1280$\times$720 pixels, and RGB frames captured by a global shutter RGB Flir camera with a resolution of 4096$\times$2196 pixels. Both of these data are then post-processed to have the same spatial resolution of 970$\times$625 pixels. Most of the sequences are short and captured with a static camera observing fast motions in the scene. Since events are confined to small regions where motion is observed, reconstructing intensity frames for other parts of the scene is not feasible. There are a few sequences recorded with a handheld camera where every pixel generates many events. We evaluate the models on ten of these handheld sequences. 

\vspace{0.25cm}
\noindent\textbf{High Speed and HDR Datasets}
These high-speed and HDR sequences are recorded by Rebecq \etal~\cite{rebecq2019high}, using a Samsung DVS Gen3 event camera \cite{son20174} with a spatial resolution of 640$\times$480. We use all three HDR sequences from this dataset, namely the $hdr\_selfie$, $hdr\_sun$, and $hdr\_tunnel$ sequences.

\section{Computational Complexity} \label{sec:computational}

\begin{table}[!t]
\small
\begin{minipage}[t]{\columnwidth}
\centering
\setlength{\tabcolsep}{2mm}
\renewcommand{\arraystretch}{1.1}
\begin{tabular}{lrcc}
\hline
    \multicolumn{1}{p{2cm}}{Network \newline Architecture}  && \multicolumn{1}{p{1.5cm}}{Number of Params (M)} & \multicolumn{1}{p{1.4cm}}{Inference Time (ms)} \\ 
    \midrule
    E2VID~\cite{rebecq2019high,stoffregen2020reducing,paredes2021back} && 
    10.71	& ~5.1 \\
    FireNet~\cite{Scheerlinck20wacv,stoffregen2020reducing}  && 
    ~0.04	& ~1.6 \\
    SPADE-E2VID~\cite{cadena2021spade} && 
    11.46	& 16.1 \\
    ET-Net~\cite{weng2021event} && 
    22.18	& 32.1 \\
    \bottomrule
\end{tabular}
\caption{\textbf{Computational complexity of different network architectures} in terms of the number of model parameters (in millions) and inference time (in milliseconds).}
\label{tab:comp}
\vspace{-0.5cm}
\end{minipage}
\end{table}

We also analyzed the computational complexity of each method by considering two metrics: the number of model parameters and inference time. The former is an essential metric as it indicates the memory requirements, while the latter reflects the real-time performance by determining the maximum FPS that can be achieved. To measure the inference time, we used a workstation equipped with a Quadro RTX 5000 GPU and considered data with a spatial resolution of 240$ \times $180. We report the average inference time for each method in ms. Table~\ref{tab:comp} compares the computational complexity of image reconstruction methods. In this table, we use the same row for the methods that share the same deep architecture. Overall, in terms of the number of parameters and inference times, FireNet is much smaller and faster than E2VID, while SPADE-E2VID is slightly larger and slower. ET-Net has the highest number of parameters which is twice as large as SPADE-E2VID, the second largest model, and its inference time is approximately 6$\times$ slower than E2VID and 20$\times$ slower than FireNet.

\section{Additional Qualitative Results} \label{sec:add_qual}

\begin{figure*}[!t]
	\newcommand{\widthplot}{0.11\textwidth}
	\centering
	\renewcommand*{\arraystretch}{1.2}
	\setlength{\tabcolsep}{0.6ex} %
\begin{tabular}{ccccccccc}
    \rotatebox[origin=l]{90}{$\;\;$boxes} &
	\includegraphics[width=\widthplot]{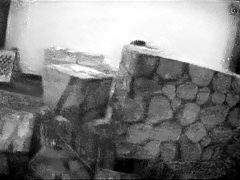} &
	\includegraphics[width=\widthplot]{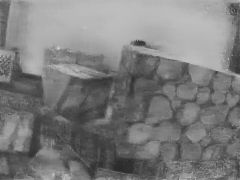} &
	\includegraphics[width=\widthplot]{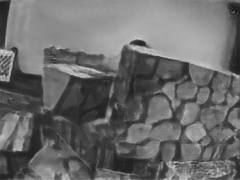} &
	\includegraphics[width=\widthplot]{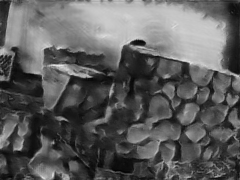} &
	\includegraphics[width=\widthplot]{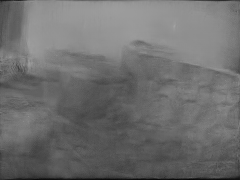} &
	\includegraphics[width=\widthplot]{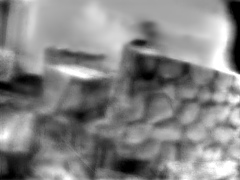} &
	\includegraphics[width=\widthplot]{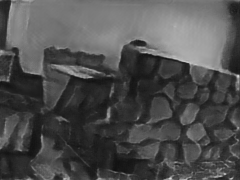} &
	\includegraphics[width=\widthplot]{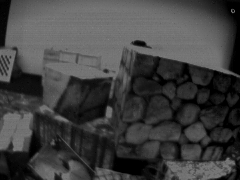} \\	
	
	\rotatebox[origin=l]{90}{$\;$calib.} &
	\includegraphics[width=\widthplot]{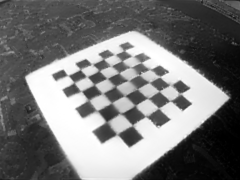} &
	\includegraphics[width=\widthplot]{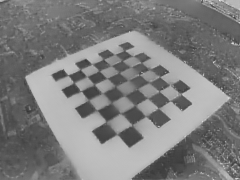} &
	\includegraphics[width=\widthplot]{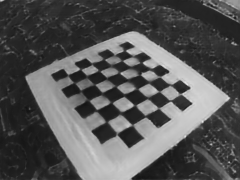} &
	\includegraphics[width=\widthplot]{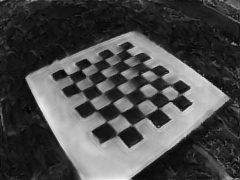} &
	\includegraphics[width=\widthplot]{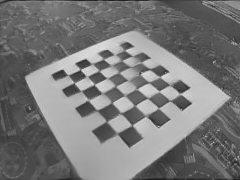} &
	\includegraphics[width=\widthplot]{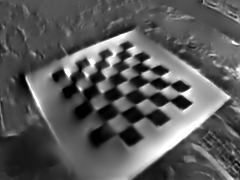} &
	\includegraphics[width=\widthplot]{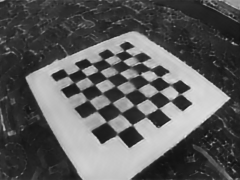} &
	\includegraphics[width=\widthplot]{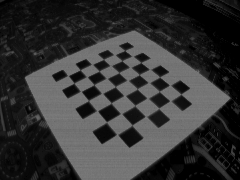} \\

	\rotatebox[origin=l]{90}{dynamic} &
	\includegraphics[width=\widthplot]{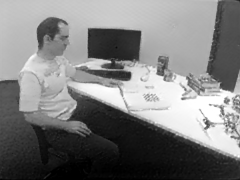} &
	\includegraphics[width=\widthplot]{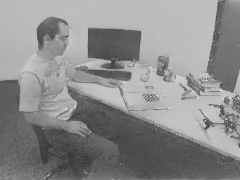} &
	\includegraphics[width=\widthplot]{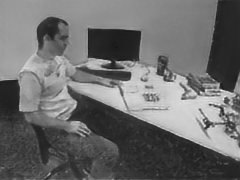} &
	\includegraphics[width=\widthplot]{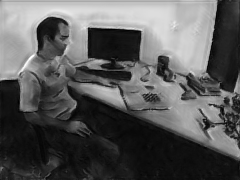} &
	\includegraphics[width=\widthplot]{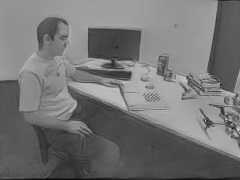} &
	\includegraphics[width=\widthplot]{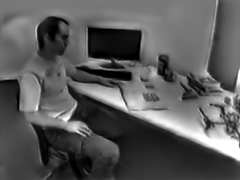} &
	\includegraphics[width=\widthplot]{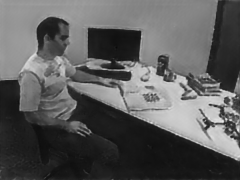} &
	\includegraphics[width=\widthplot]{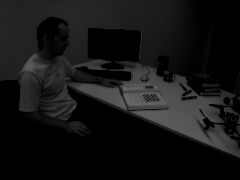} \\

	\rotatebox[origin=l]{90}{$\;\;$office} &
	\includegraphics[width=\widthplot]{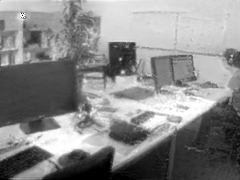} &
	\includegraphics[width=\widthplot]{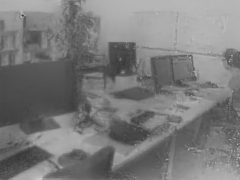} &
	\includegraphics[width=\widthplot]{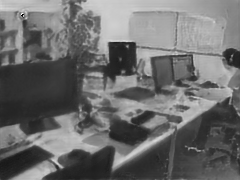} &
	\includegraphics[width=\widthplot]{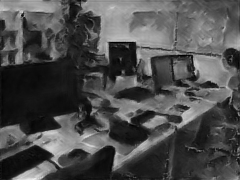} &
	\includegraphics[width=\widthplot]{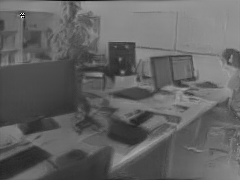} &
	\includegraphics[width=\widthplot]{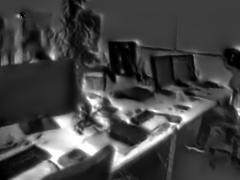} &
	\includegraphics[width=\widthplot]{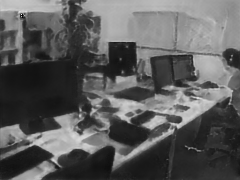} &
	\includegraphics[width=\widthplot]{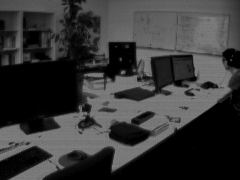} \\

	\rotatebox[origin=l]{90}{$\;\;$poster} &
	\includegraphics[width=\widthplot]{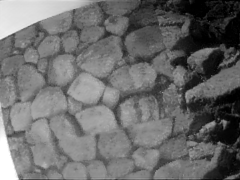} &
	\includegraphics[width=\widthplot]{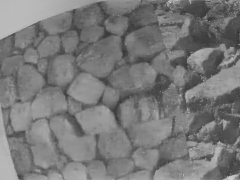} &
	\includegraphics[width=\widthplot]{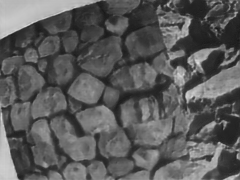} &
	\includegraphics[width=\widthplot]{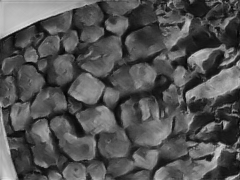} &
	\includegraphics[width=\widthplot]{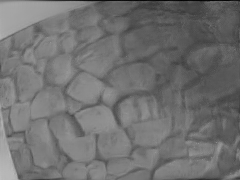} &
	\includegraphics[width=\widthplot]{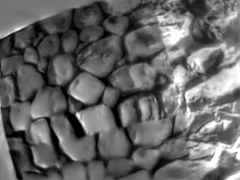} &
	\includegraphics[width=\widthplot]{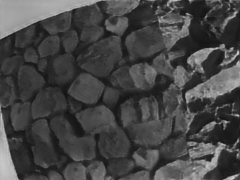} &
	\includegraphics[width=\widthplot]{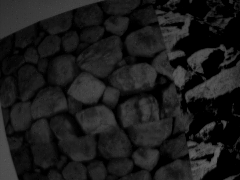} \\
 
	\rotatebox[origin=l]{90}{$\;\;$slider} &
	\includegraphics[width=\widthplot]{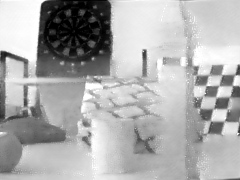} &
	\includegraphics[width=\widthplot]{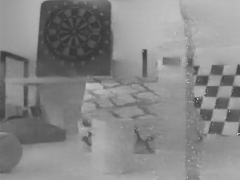} &
	\includegraphics[width=\widthplot]{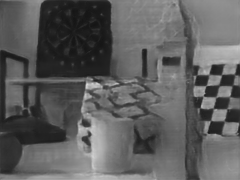} &
	\includegraphics[width=\widthplot]{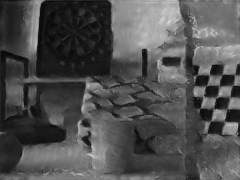} &
	\includegraphics[width=\widthplot]{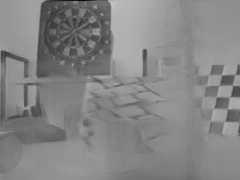} &
	\includegraphics[width=\widthplot]{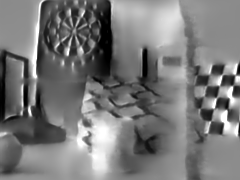} &
	\includegraphics[width=\widthplot]{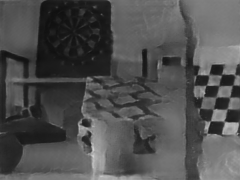} &
	\includegraphics[width=\widthplot]{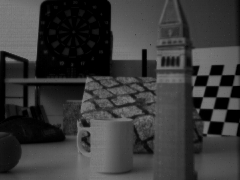} \\

	& E2VID & FireNet & E2VID+ & FireNet+ & \makebox[0pt][c]{SPADE-E2VID} & SSL-E2VID & ET-Net & Ground Truth 

\end{tabular}
	\caption{Additional qualitative comparisons on the ECD dataset.}
	\label{fig:qual_eval_ECD}
        \vspace{0.2cm}
\end{figure*}

\begin{figure*}[!t]
	\newcommand{\widthplot}{0.11\textwidth}
	\centering
	\renewcommand*{\arraystretch}{1.2}
	\setlength{\tabcolsep}{0.6ex} %
\begin{tabular}{ccccccccc}
    \rotatebox[origin=l]{90}{indoor1} &
	\includegraphics[width=\widthplot]{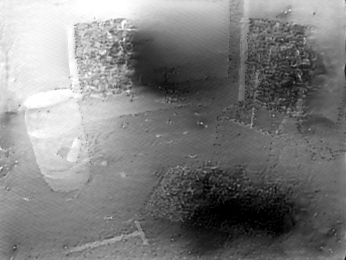} &
	\includegraphics[width=\widthplot]{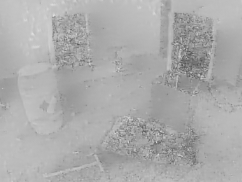} &
	\includegraphics[width=\widthplot]{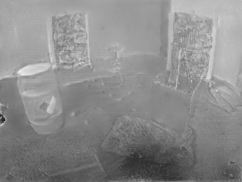} &
	\includegraphics[width=\widthplot]{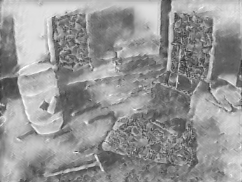} &
	\includegraphics[width=\widthplot]{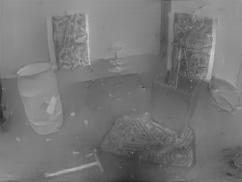} &
	\includegraphics[width=\widthplot]{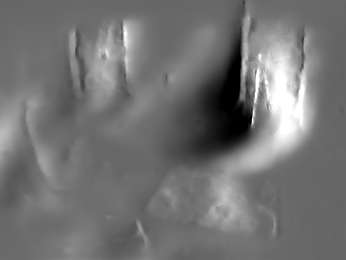} &
	\includegraphics[width=\widthplot]{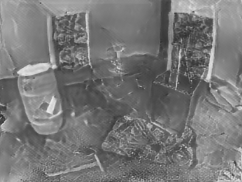} &
	\includegraphics[width=\widthplot]{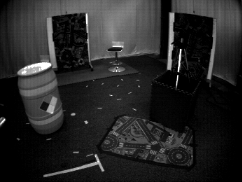} \\	
	
	\rotatebox[origin=l]{90}{indoor2} &
	\includegraphics[width=\widthplot]{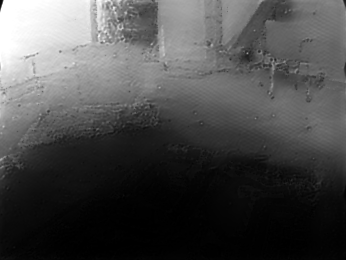} &
	\includegraphics[width=\widthplot]{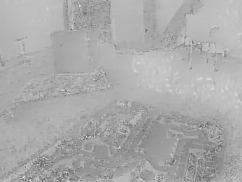} &
	\includegraphics[width=\widthplot]{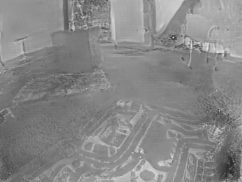} &
	\includegraphics[width=\widthplot]{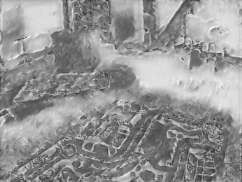} &
	\includegraphics[width=\widthplot]{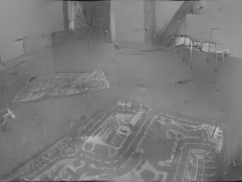} &
	\includegraphics[width=\widthplot]{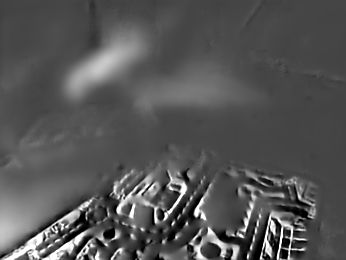} &
	\includegraphics[width=\widthplot]{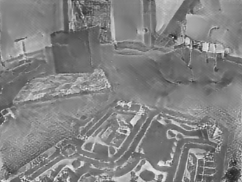} &
	\includegraphics[width=\widthplot]{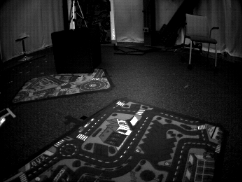} \\

	\rotatebox[origin=l]{90}{indoor4} &
	\includegraphics[width=\widthplot]{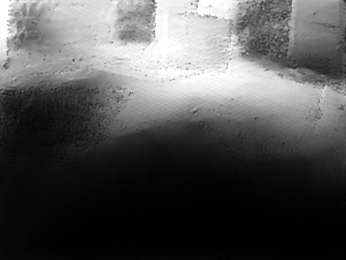} &
	\includegraphics[width=\widthplot]{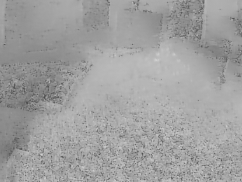} &
	\includegraphics[width=\widthplot]{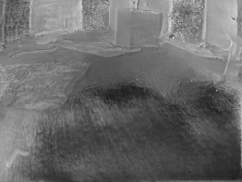} &
	\includegraphics[width=\widthplot]{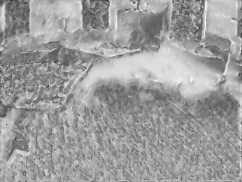} &
	\includegraphics[width=\widthplot]{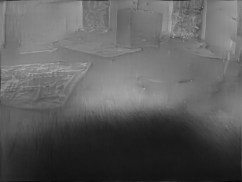} &
	\includegraphics[width=\widthplot]{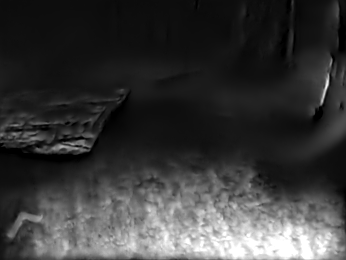} &
	\includegraphics[width=\widthplot]{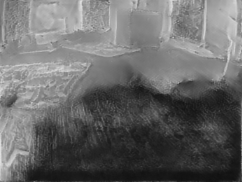} &
	\includegraphics[width=\widthplot]{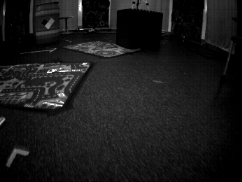} \\

	\rotatebox[origin=l]{90}{outdoor2} &
	\includegraphics[width=\widthplot]{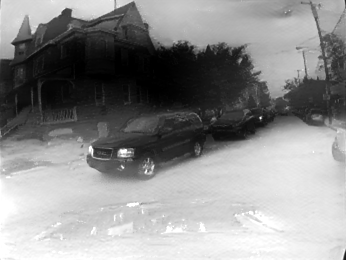} &
	\includegraphics[width=\widthplot]{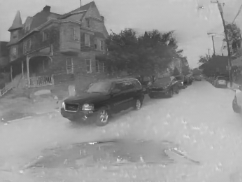} &
	\includegraphics[width=\widthplot]{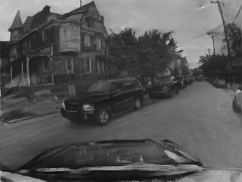} &
	\includegraphics[width=\widthplot]{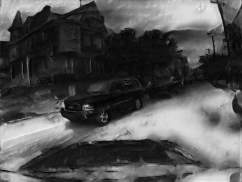} &
	\includegraphics[width=\widthplot]{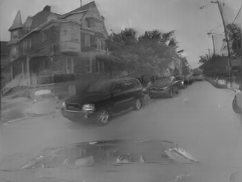} &
	\includegraphics[width=\widthplot]{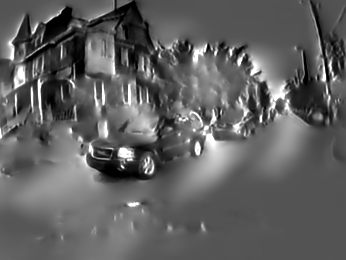} &
	\includegraphics[width=\widthplot]{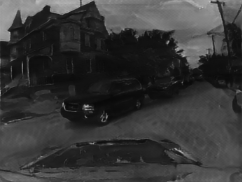} &
	\includegraphics[width=\widthplot]{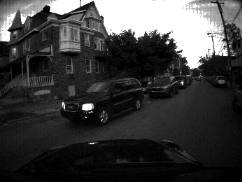} \\

	\rotatebox[origin=l]{90}{outdoor2} &
	\includegraphics[width=\widthplot]{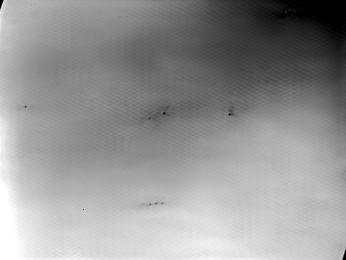} &
	\includegraphics[width=\widthplot]{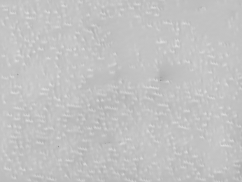} &
	\includegraphics[width=\widthplot]{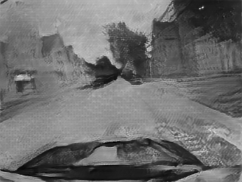} &
	\includegraphics[width=\widthplot]{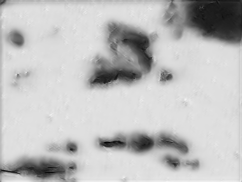} &
	\includegraphics[width=\widthplot]{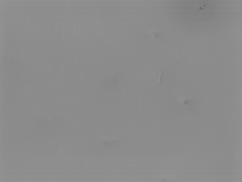} &
	\includegraphics[width=\widthplot]{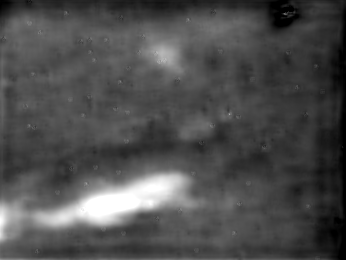} &
	\includegraphics[width=\widthplot]{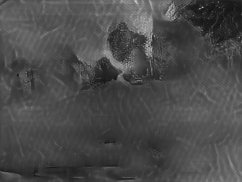} &
	\includegraphics[width=\widthplot]{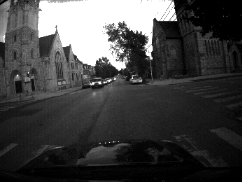} \\
 
	\rotatebox[origin=l]{90}{outdoor2} &
	\includegraphics[width=\widthplot]{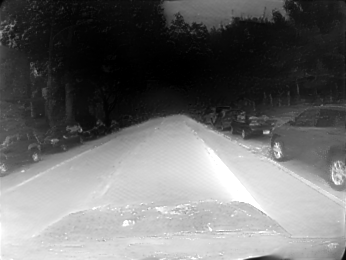} &
	\includegraphics[width=\widthplot]{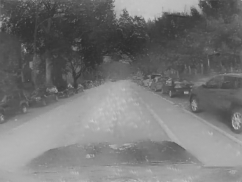} &
	\includegraphics[width=\widthplot]{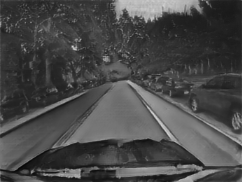} &
	\includegraphics[width=\widthplot]{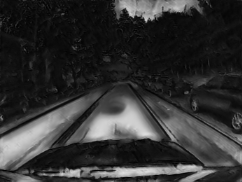} &
	\includegraphics[width=\widthplot]{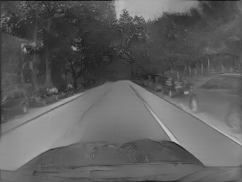} &
	\includegraphics[width=\widthplot]{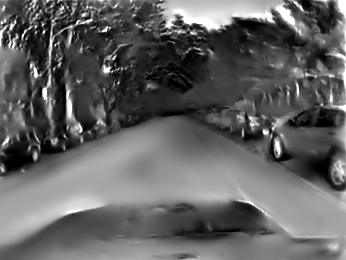} &
	\includegraphics[width=\widthplot]{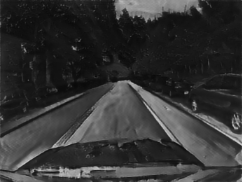} &
	\includegraphics[width=\widthplot]{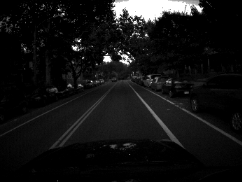} \\

	& E2VID & FireNet & E2VID+ & FireNet+ & \makebox[0pt][c]{SPADE-E2VID} & SSL-E2VID & ET-Net & Ground Truth 

\end{tabular}
	\caption{Additional qualitative comparisons on the MVSEC dataset.}
	\label{fig:qual_eval_MVSEC}
\end{figure*}

\begin{figure*}[!t]
	\newcommand{\widthplot}{0.11\textwidth}
	\centering
	\renewcommand*{\arraystretch}{1.2}
	\setlength{\tabcolsep}{0.6ex} %
\begin{tabular}{ccccccccc}
    \rotatebox[origin=l]{90}{bike\_bay} &
	\includegraphics[width=\widthplot]{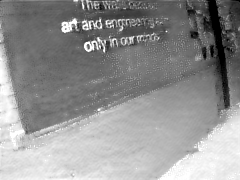} &
	\includegraphics[width=\widthplot]{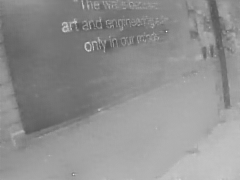} &
	\includegraphics[width=\widthplot]{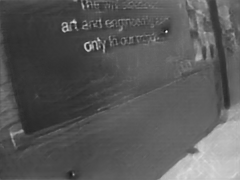} &
	\includegraphics[width=\widthplot]{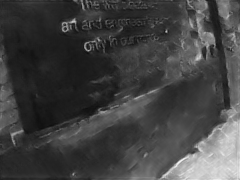} &
	\includegraphics[width=\widthplot]{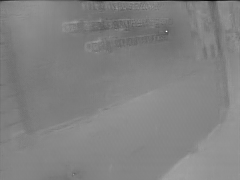} &
	\includegraphics[width=\widthplot]{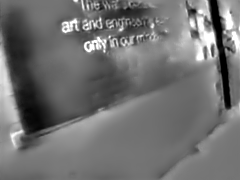} &
	\includegraphics[width=\widthplot]{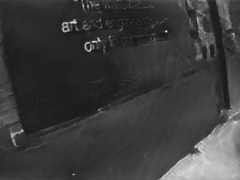} &
	\includegraphics[width=\widthplot]{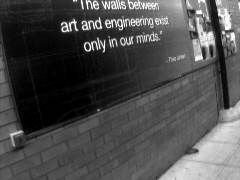} \\	
	
	\rotatebox[origin=l]{90}{$\;$boxes} &
	\includegraphics[width=\widthplot]{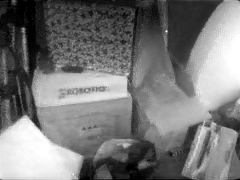} &
	\includegraphics[width=\widthplot]{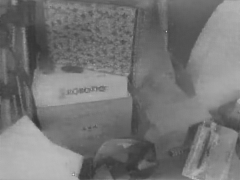} &
	\includegraphics[width=\widthplot]{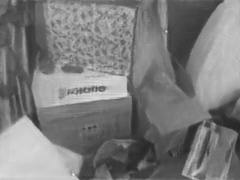} &
	\includegraphics[width=\widthplot]{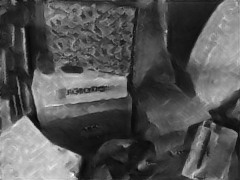} &
	\includegraphics[width=\widthplot]{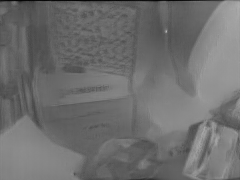} &
	\includegraphics[width=\widthplot]{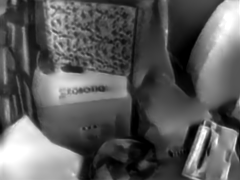} &
	\includegraphics[width=\widthplot]{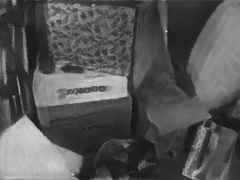} &
	\includegraphics[width=\widthplot]{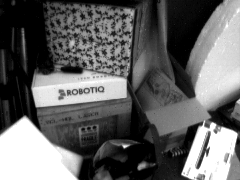} \\

	\rotatebox[origin=l]{90}{plants} &
	\includegraphics[width=\widthplot]{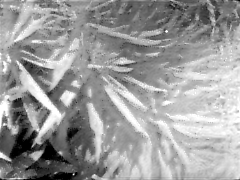} &
	\includegraphics[width=\widthplot]{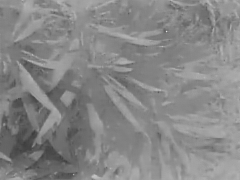} &
	\includegraphics[width=\widthplot]{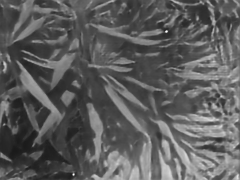} &
	\includegraphics[width=\widthplot]{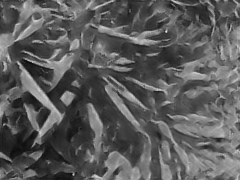} &
	\includegraphics[width=\widthplot]{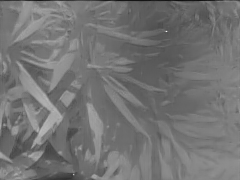} &
	\includegraphics[width=\widthplot]{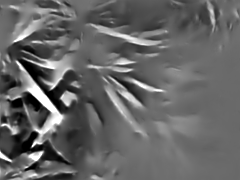} &
	\includegraphics[width=\widthplot]{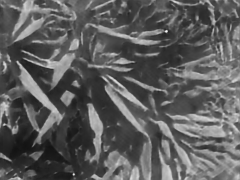} &
	\includegraphics[width=\widthplot]{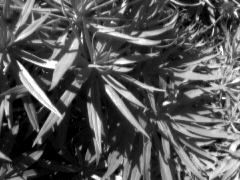} \\

	\rotatebox[origin=l]{90}{$\;\;$poster} &
	\includegraphics[width=\widthplot]{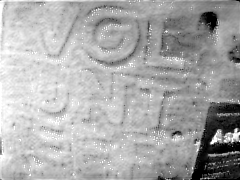} &
	\includegraphics[width=\widthplot]{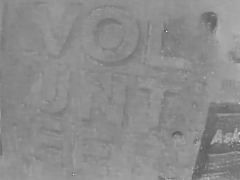} &
	\includegraphics[width=\widthplot]{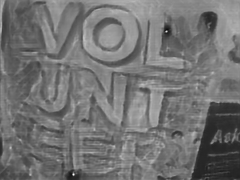} &
	\includegraphics[width=\widthplot]{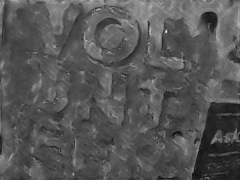} &
	\includegraphics[width=\widthplot]{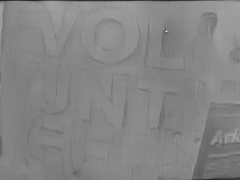} &
	\includegraphics[width=\widthplot]{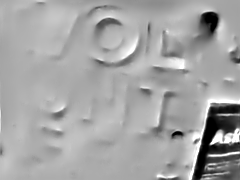} &
	\includegraphics[width=\widthplot]{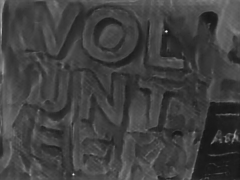} &
	\includegraphics[width=\widthplot]{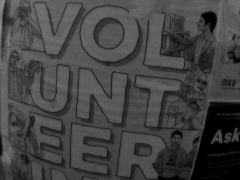} \\

	\rotatebox[origin=l]{90}{reflect.} &
	\includegraphics[width=\widthplot]{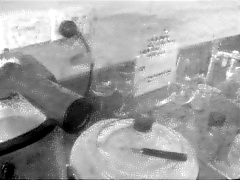} &
	\includegraphics[width=\widthplot]{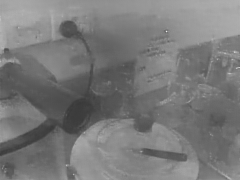} &
	\includegraphics[width=\widthplot]{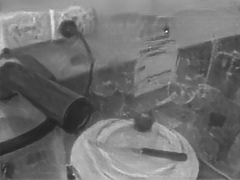} &
	\includegraphics[width=\widthplot]{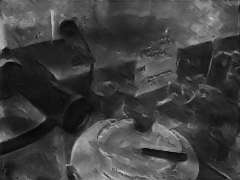} &
	\includegraphics[width=\widthplot]{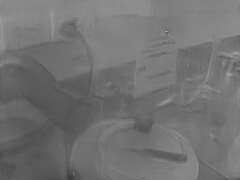} &
	\includegraphics[width=\widthplot]{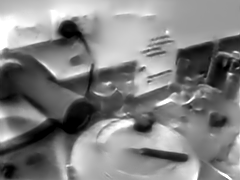} &
	\includegraphics[width=\widthplot]{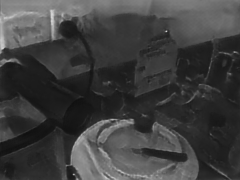} &
	\includegraphics[width=\widthplot]{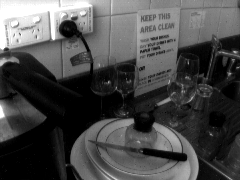} \\
 
	\rotatebox[origin=l]{90}{slowhand} &
	\includegraphics[width=\widthplot]{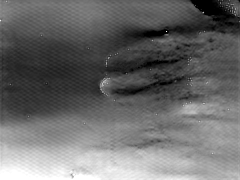} &
	\includegraphics[width=\widthplot]{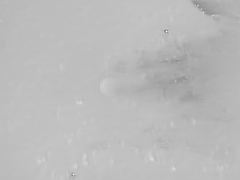} &
	\includegraphics[width=\widthplot]{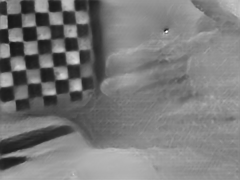} &
	\includegraphics[width=\widthplot]{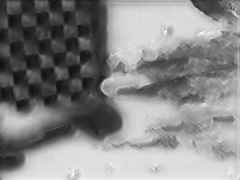} &
	\includegraphics[width=\widthplot]{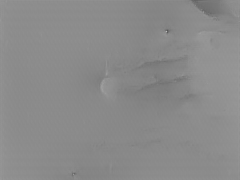} &
	\includegraphics[width=\widthplot]{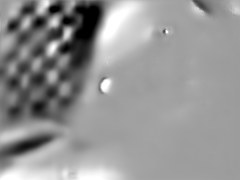} &
	\includegraphics[width=\widthplot]{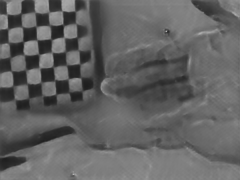} &
	\includegraphics[width=\widthplot]{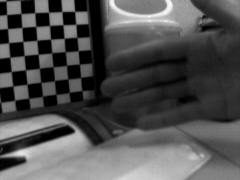} \\

	\rotatebox[origin=l]{90}{still\_life} &
	\includegraphics[width=\widthplot]{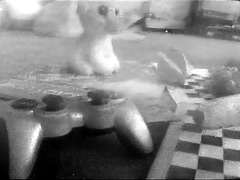} &
	\includegraphics[width=\widthplot]{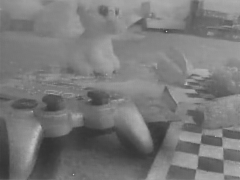} &
	\includegraphics[width=\widthplot]{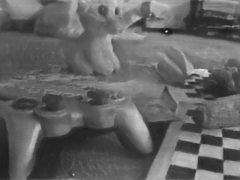} &
	\includegraphics[width=\widthplot]{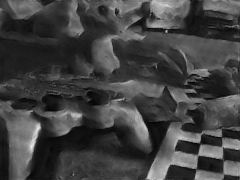} &
	\includegraphics[width=\widthplot]{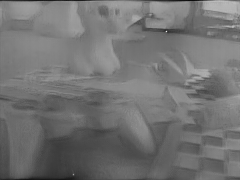} &
	\includegraphics[width=\widthplot]{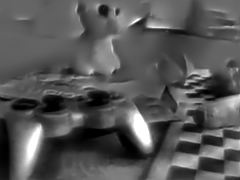} &
	\includegraphics[width=\widthplot]{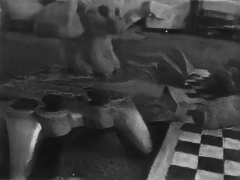} &
	\includegraphics[width=\widthplot]{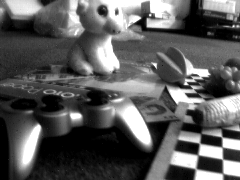} \\

	& E2VID & FireNet & E2VID+ & FireNet+ & \makebox[0pt][c]{SPADE-E2VID} & SSL-E2VID & ET-Net & Ground Truth 

\end{tabular}
	\caption{Additional qualitative comparisons on the HQF dataset.}
	\label{fig:qual_eval_HQF}
        \vspace{1cm}
\end{figure*}

\begin{figure*}[!t]
	\newcommand{\widthplot}{0.11\textwidth}
	\centering
	\renewcommand*{\arraystretch}{1.2}
	\setlength{\tabcolsep}{0.6ex} %
\begin{tabular}{ccccccccc}
    \rotatebox[origin=l]{90}{handh.1} &
	\includegraphics[width=\widthplot]{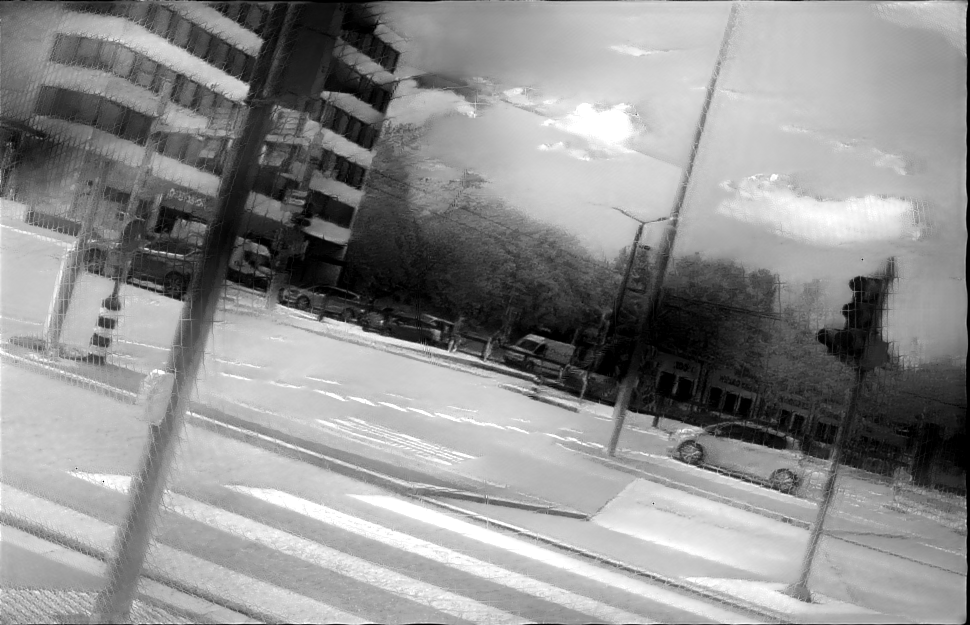} &
	\includegraphics[width=\widthplot]{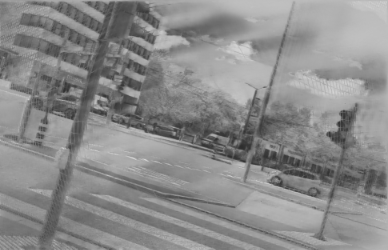} &
	\includegraphics[width=\widthplot]{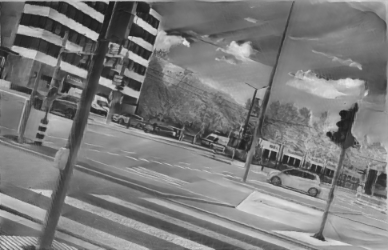} &
	\includegraphics[width=\widthplot]{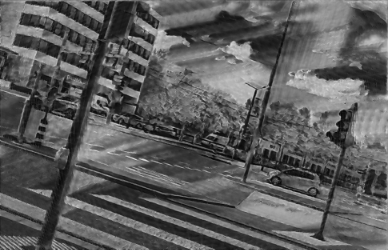} &
	\includegraphics[width=\widthplot]{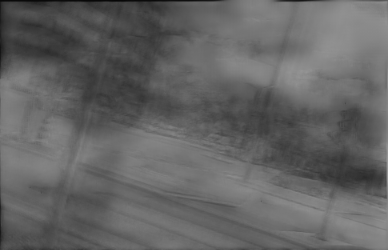} &
	\includegraphics[width=\widthplot]{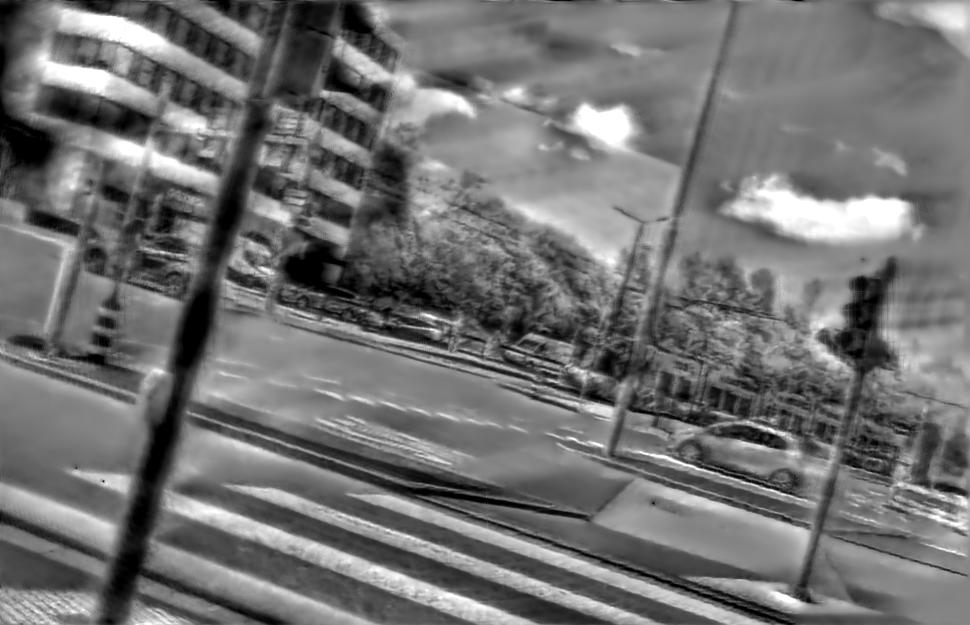} &
	\includegraphics[width=\widthplot]{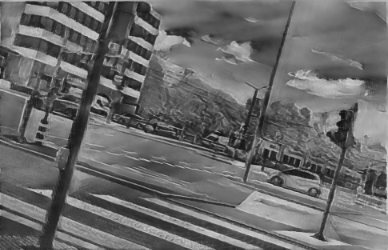} &
	\includegraphics[width=\widthplot]{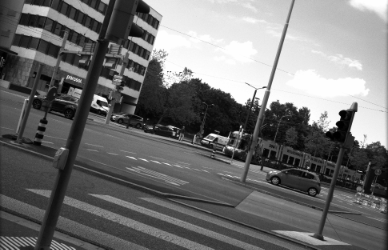} \\	
	
	\rotatebox[origin=l]{90}{handh.3} &
	\includegraphics[width=\widthplot]{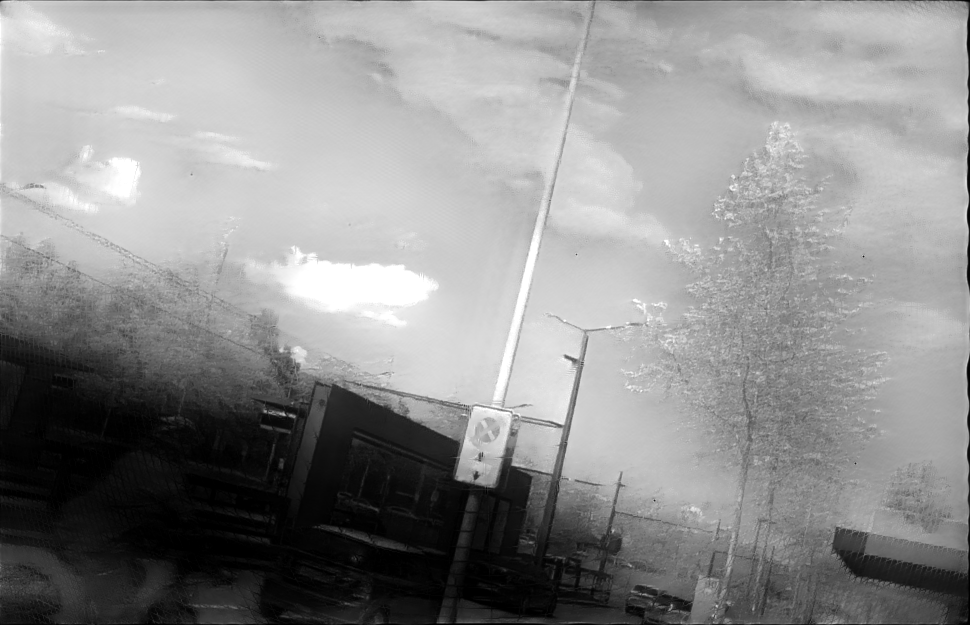} &
	\includegraphics[width=\widthplot]{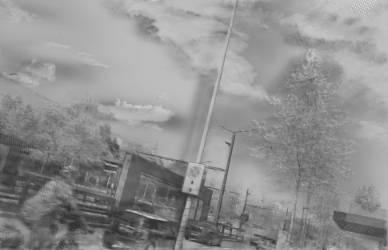} &
	\includegraphics[width=\widthplot]{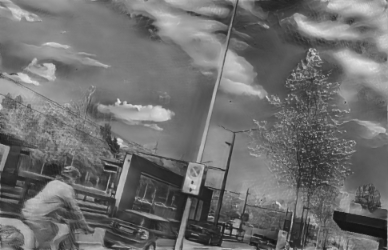} &
	\includegraphics[width=\widthplot]{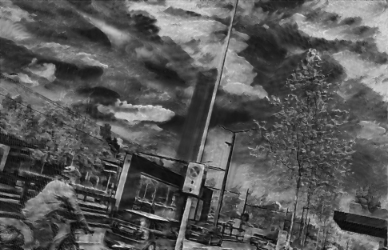} &
	\includegraphics[width=\widthplot]{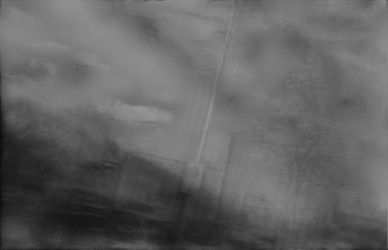} &
	\includegraphics[width=\widthplot]{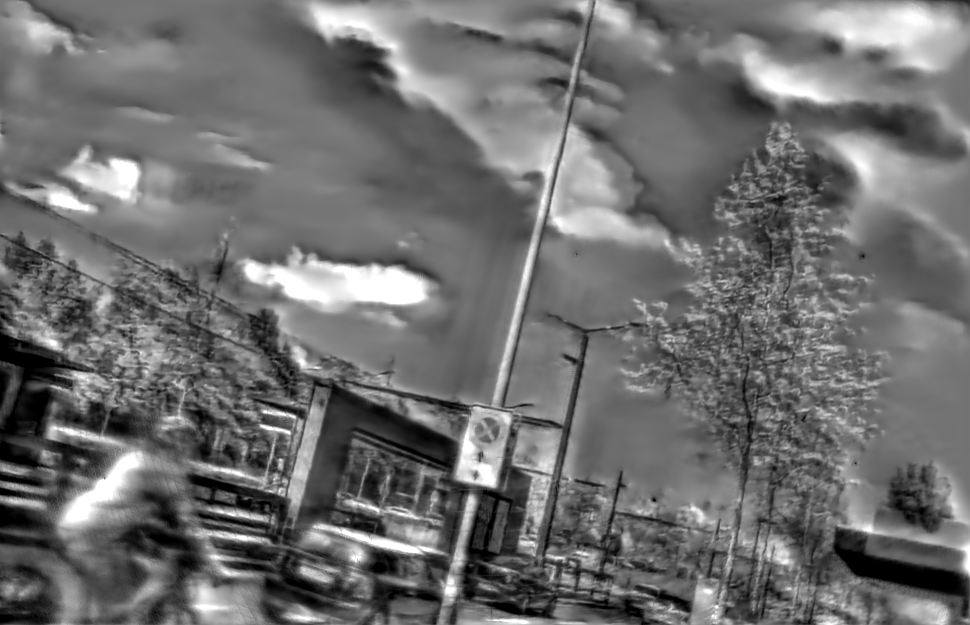} &
	\includegraphics[width=\widthplot]{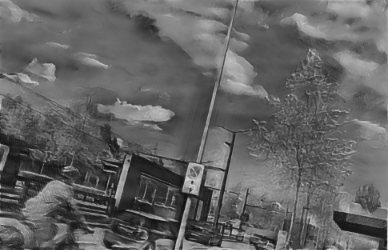} &
	\includegraphics[width=\widthplot]{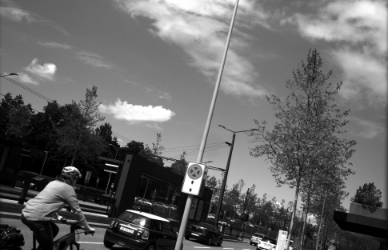} \\

	\rotatebox[origin=l]{90}{rooftop1} &
	\includegraphics[width=\widthplot]{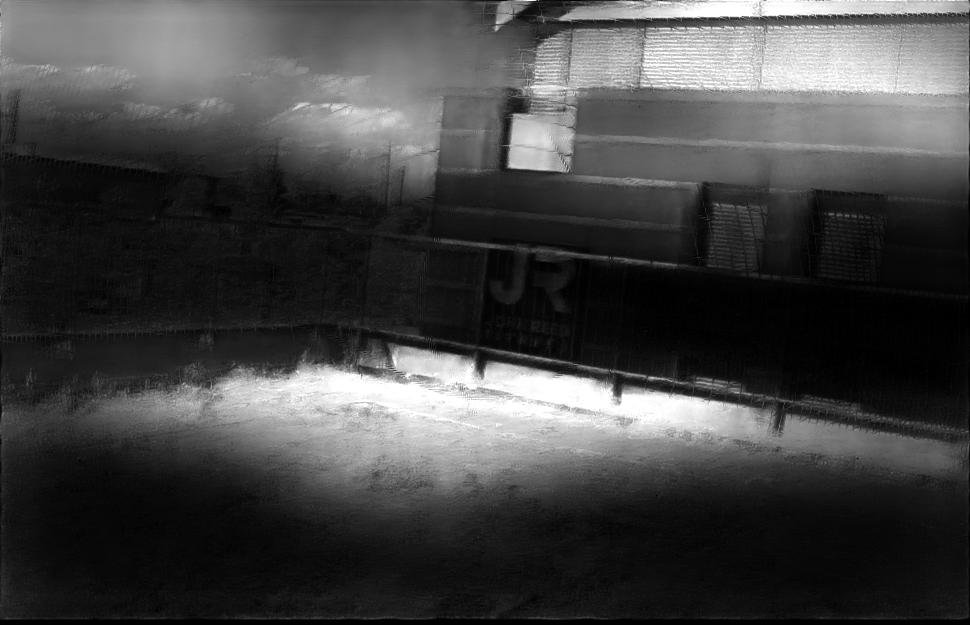} &
	\includegraphics[width=\widthplot]{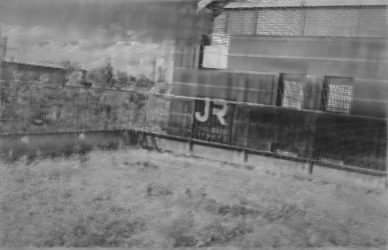} &
	\includegraphics[width=\widthplot]{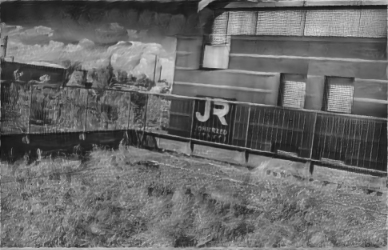} &
	\includegraphics[width=\widthplot]{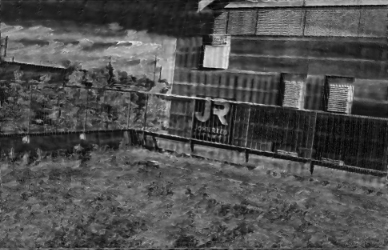} &
	\includegraphics[width=\widthplot]{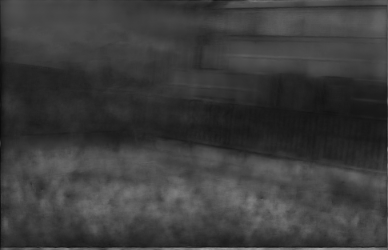} &
	\includegraphics[width=\widthplot]{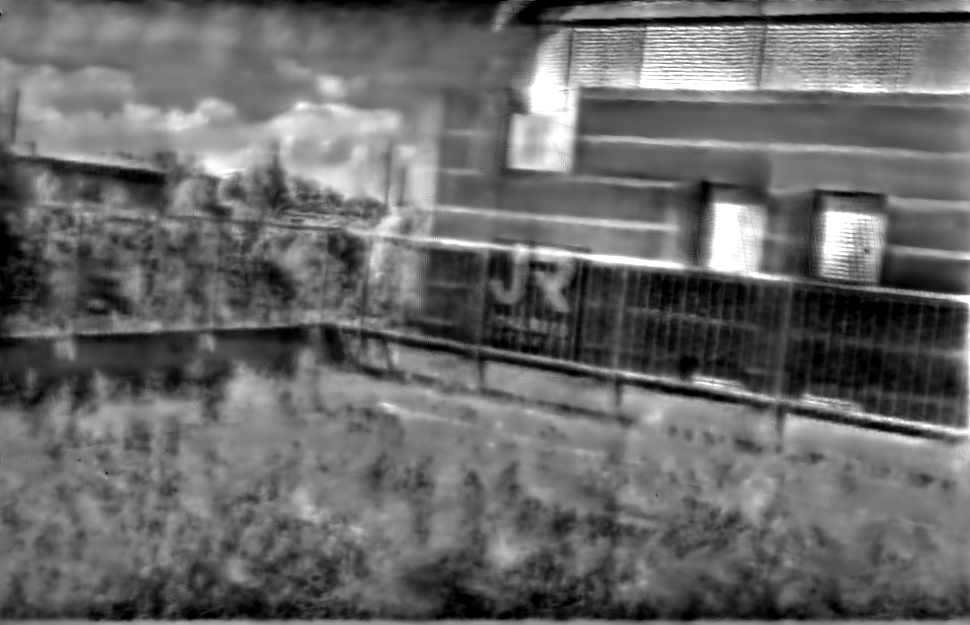} &
	\includegraphics[width=\widthplot]{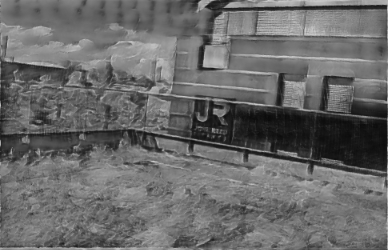} &
	\includegraphics[width=\widthplot]{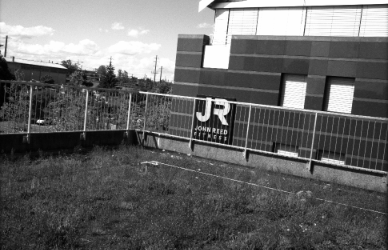} \\

	\rotatebox[origin=l]{90}{rooftop2} &
	\includegraphics[width=\widthplot]{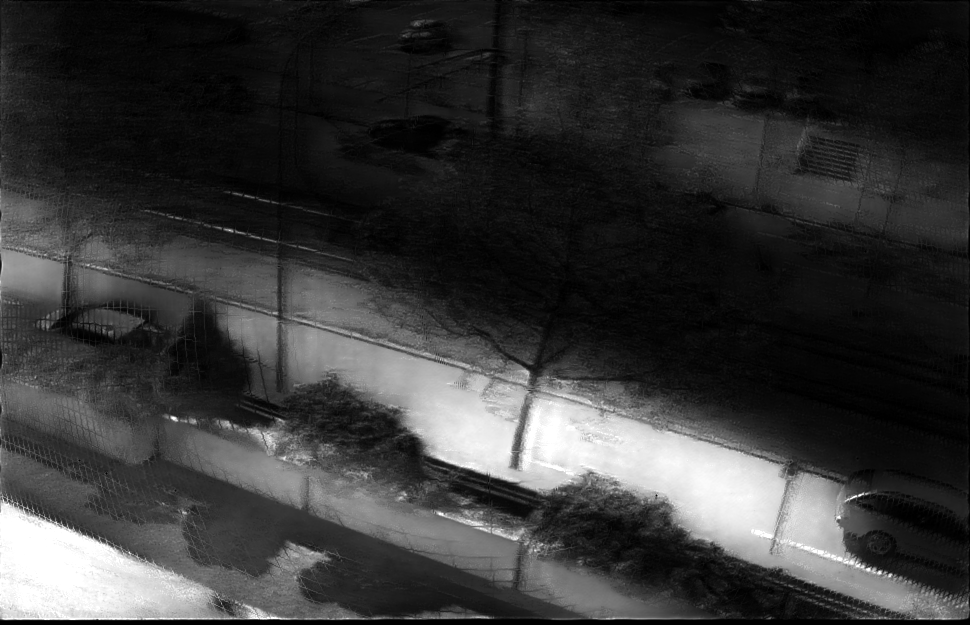} &
	\includegraphics[width=\widthplot]{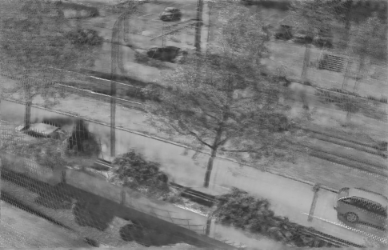} &
	\includegraphics[width=\widthplot]{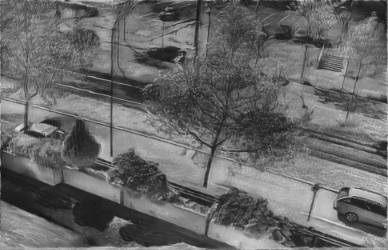} &
	\includegraphics[width=\widthplot]{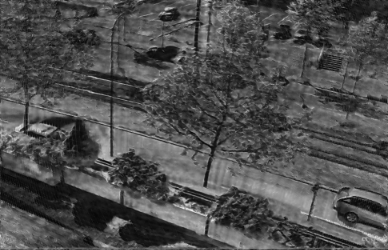} &
	\includegraphics[width=\widthplot]{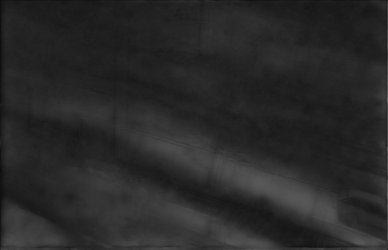} &
	\includegraphics[width=\widthplot]{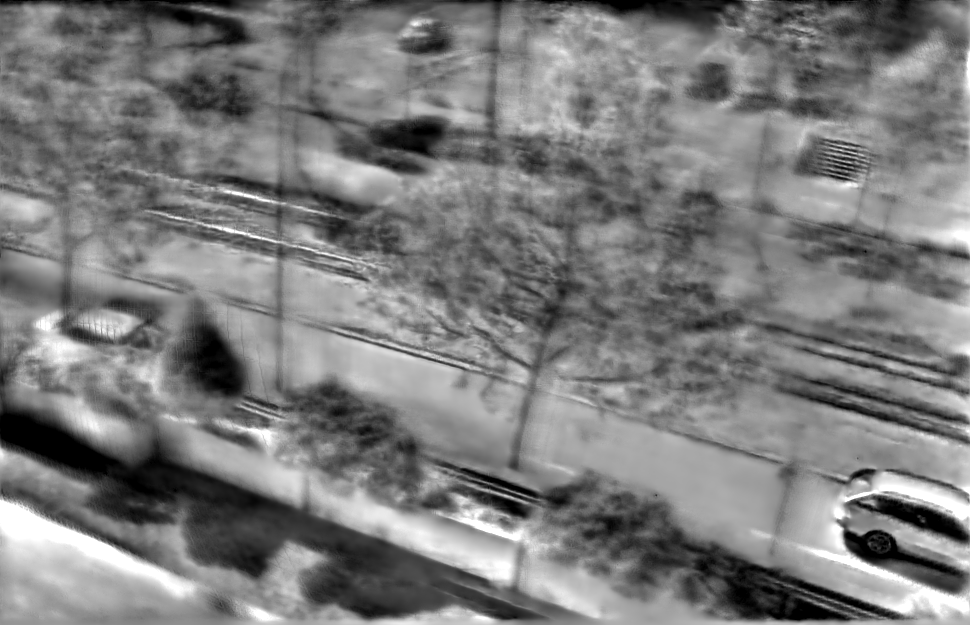} &
	\includegraphics[width=\widthplot]{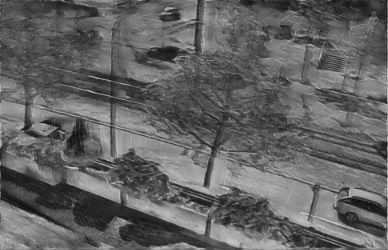} &
	\includegraphics[width=\widthplot]{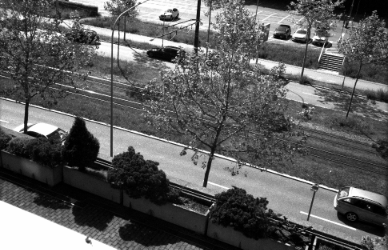} \\

	\rotatebox[origin=l]{90}{$\;$street} &
	\includegraphics[width=\widthplot]{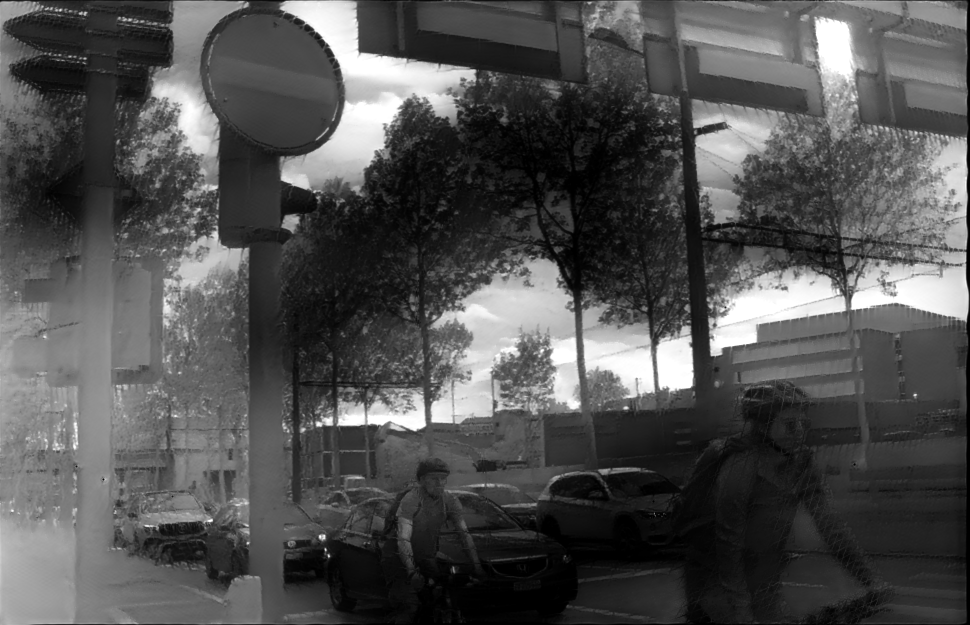} &
	\includegraphics[width=\widthplot]{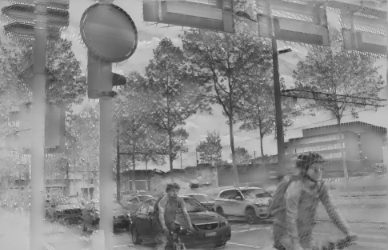} &
	\includegraphics[width=\widthplot]{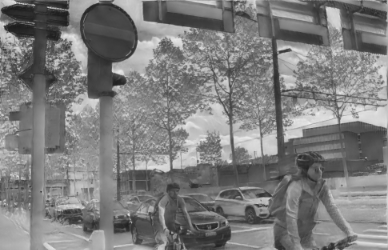} &
	\includegraphics[width=\widthplot]{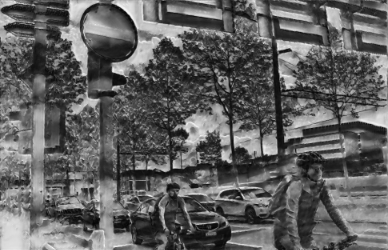} &
	\includegraphics[width=\widthplot]{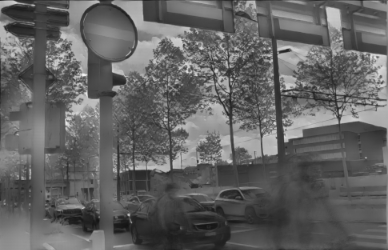} &
	\includegraphics[width=\widthplot]{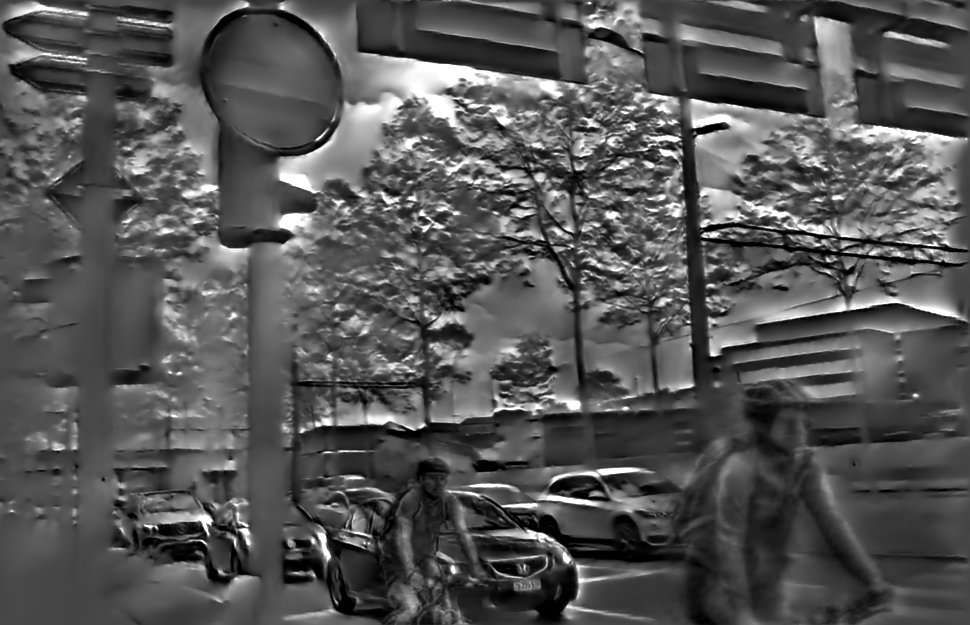} &
	\includegraphics[width=\widthplot]{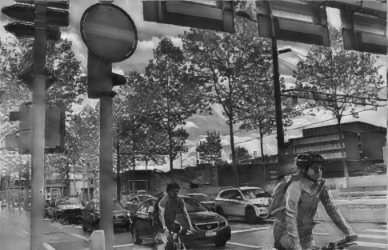} &
	\includegraphics[width=\widthplot]{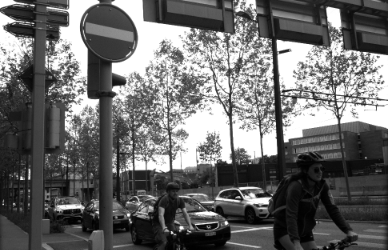} \\
 
	& E2VID & FireNet & E2VID+ & FireNet+ & \makebox[0pt][c]{SPADE-E2VID} & SSL-E2VID & ET-Net & Ground Truth 

\end{tabular}
	\caption{Additional qualitative comparisons on the BS-ERGB dataset.}
	\label{fig:qual_eval_BSERGB}
\end{figure*}

\begin{figure*}[!t]
	\newcommand{\widthplot}{0.11\textwidth}
	\centering
	\renewcommand*{\arraystretch}{1.2}
	\setlength{\tabcolsep}{0.6ex} %
\begin{tabular}{ccccccccc}
    \rotatebox[origin=l]{90}{$\;\;$calib.} &
	\includegraphics[width=\widthplot]{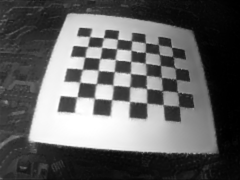} &
	\includegraphics[width=\widthplot]{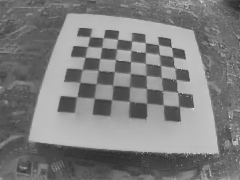} &
	\includegraphics[width=\widthplot]{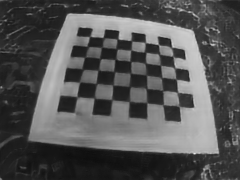} &
	\includegraphics[width=\widthplot]{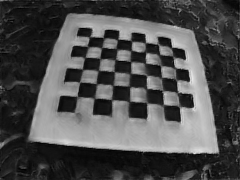} &
	\includegraphics[width=\widthplot]{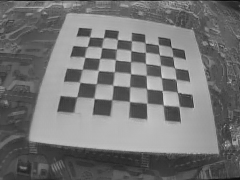} &
	\includegraphics[width=\widthplot]{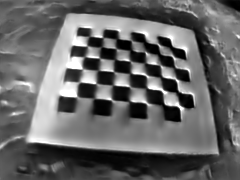} &
	\includegraphics[width=\widthplot]{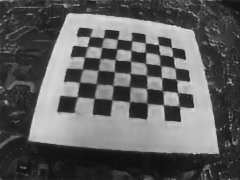} &
	\includegraphics[width=\widthplot]{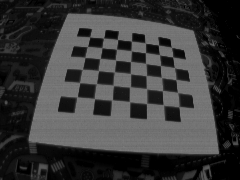} \\	
	
	\rotatebox[origin=l]{90}{dynamic} &
	\includegraphics[width=\widthplot]{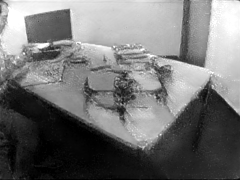} &
	\includegraphics[width=\widthplot]{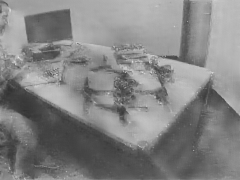} &
	\includegraphics[width=\widthplot]{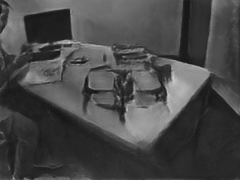} &
	\includegraphics[width=\widthplot]{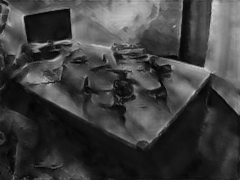} &
	\includegraphics[width=\widthplot]{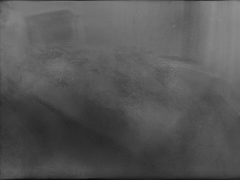} &
	\includegraphics[width=\widthplot]{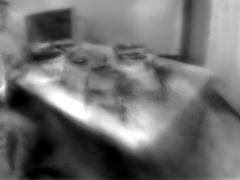} &
	\includegraphics[width=\widthplot]{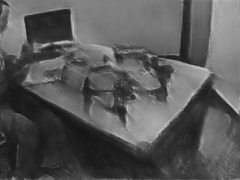} &
	\includegraphics[width=\widthplot]{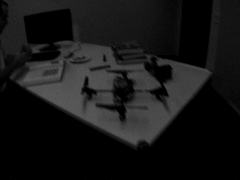} \\

	\rotatebox[origin=l]{90}{$\;\;$poster} &
	\includegraphics[width=\widthplot]{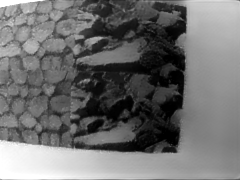} &
	\includegraphics[width=\widthplot]{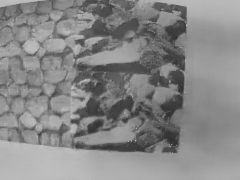} &
	\includegraphics[width=\widthplot]{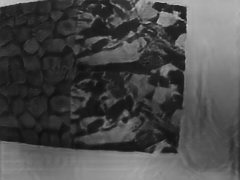} &
	\includegraphics[width=\widthplot]{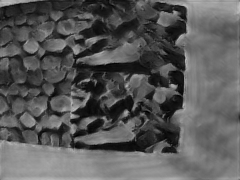} &
	\includegraphics[width=\widthplot]{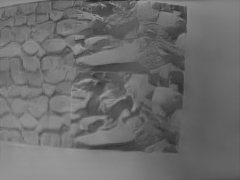} &
	\includegraphics[width=\widthplot]{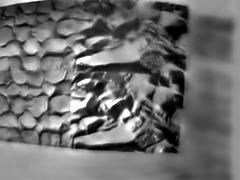} &
	\includegraphics[width=\widthplot]{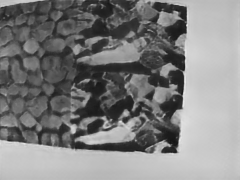} &
	\includegraphics[width=\widthplot]{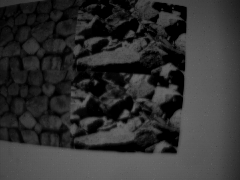} \\

	\rotatebox[origin=l]{90}{$\;\;$shapes} &
	\includegraphics[width=\widthplot]{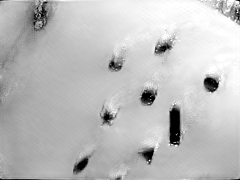} &
	\includegraphics[width=\widthplot]{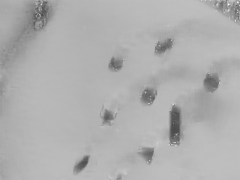} &
	\includegraphics[width=\widthplot]{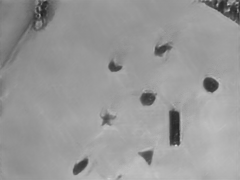} &
	\includegraphics[width=\widthplot]{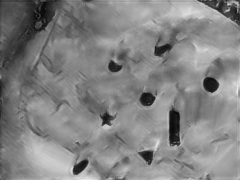} &
	\includegraphics[width=\widthplot]{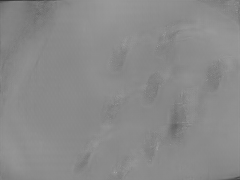} &
	\includegraphics[width=\widthplot]{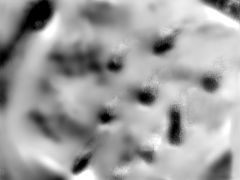} &
	\includegraphics[width=\widthplot]{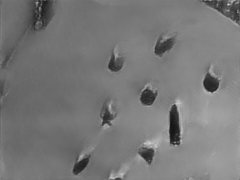} &
	\includegraphics[width=\widthplot]{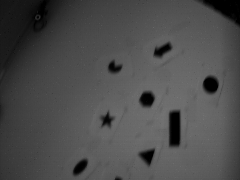} \\

	& E2VID & FireNet & E2VID+ & FireNet+ & \makebox[0pt][c]{SPADE-E2VID} & SSL-E2VID & ET-Net & Reference 

\end{tabular}
	\caption{Additional qualitative comparisons on the fast parts of the ECD dataset (ECD-FAST).}
	\label{fig:qual_eval_ECD_fast}
        \vspace{1cm}
\end{figure*}

\begin{figure*}[!t]
	\newcommand{\widthplot}{0.11\textwidth}
	\centering
	\renewcommand*{\arraystretch}{1.2}
	\setlength{\tabcolsep}{0.6ex} %
\begin{tabular}{ccccccccc}
    \rotatebox[origin=l]{90}{night1} &
	\includegraphics[width=\widthplot]{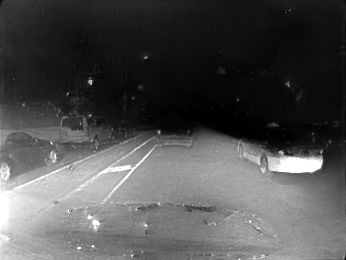} &
	\includegraphics[width=\widthplot]{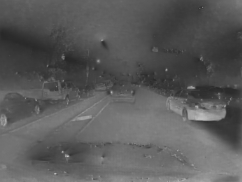} &
	\includegraphics[width=\widthplot]{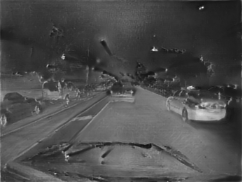} &
	\includegraphics[width=\widthplot]{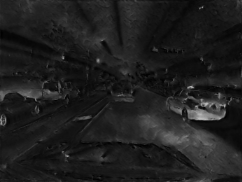} &
	\includegraphics[width=\widthplot]{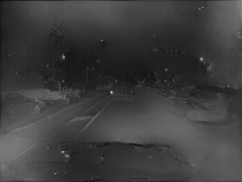} &
	\includegraphics[width=\widthplot]{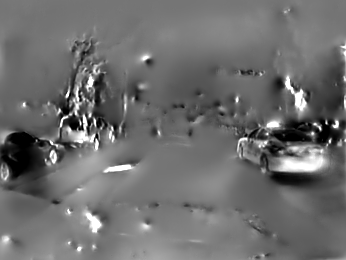} &
	\includegraphics[width=\widthplot]{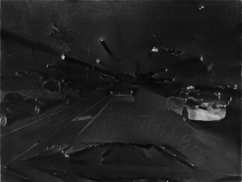} &
	\includegraphics[width=\widthplot]{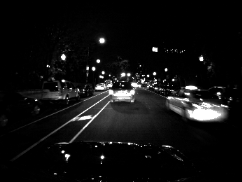} \\	
	
	\rotatebox[origin=l]{90}{night1} &
	\includegraphics[width=\widthplot]{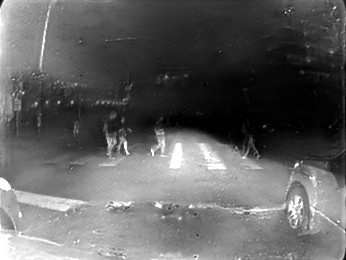} &
	\includegraphics[width=\widthplot]{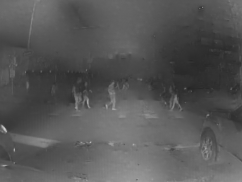} &
	\includegraphics[width=\widthplot]{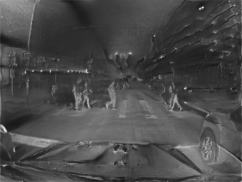} &
	\includegraphics[width=\widthplot]{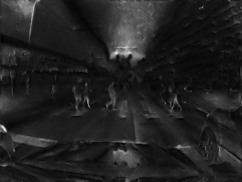} &
	\includegraphics[width=\widthplot]{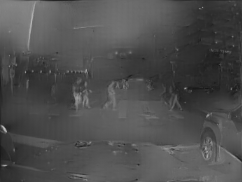} &
	\includegraphics[width=\widthplot]{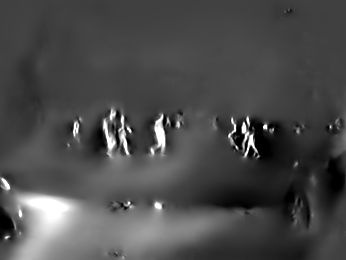} &
	\includegraphics[width=\widthplot]{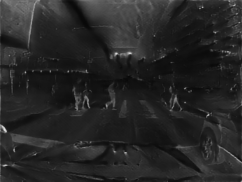} &
	\includegraphics[width=\widthplot]{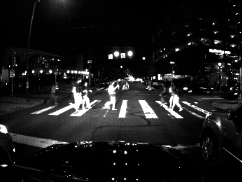} \\

	\rotatebox[origin=l]{90}{night2} &
	\includegraphics[width=\widthplot]{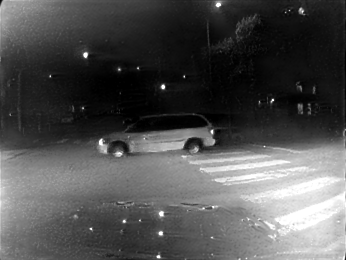} &
	\includegraphics[width=\widthplot]{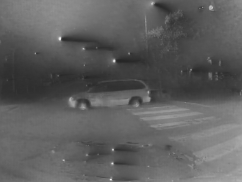} &
	\includegraphics[width=\widthplot]{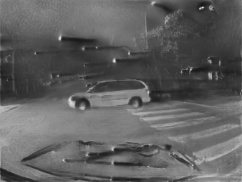} &
	\includegraphics[width=\widthplot]{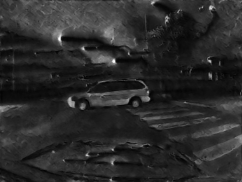} &
	\includegraphics[width=\widthplot]{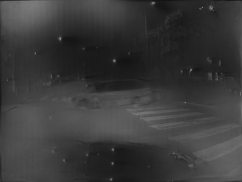} &
	\includegraphics[width=\widthplot]{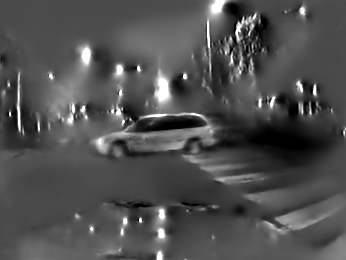} &
	\includegraphics[width=\widthplot]{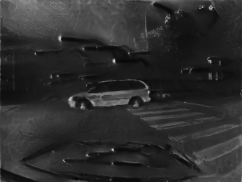} &
	\includegraphics[width=\widthplot]{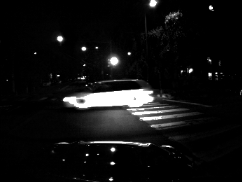} \\

	\rotatebox[origin=l]{90}{night2} &
	\includegraphics[width=\widthplot]{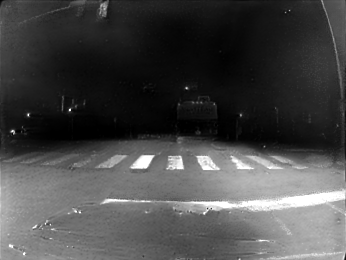} &
	\includegraphics[width=\widthplot]{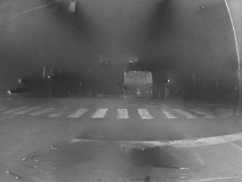} &
	\includegraphics[width=\widthplot]{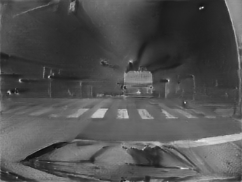} &
	\includegraphics[width=\widthplot]{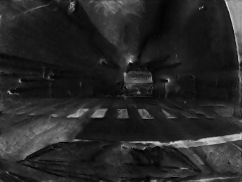} &
	\includegraphics[width=\widthplot]{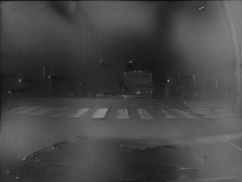} &
	\includegraphics[width=\widthplot]{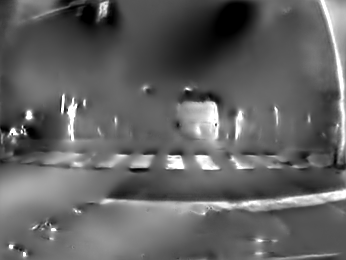} &
	\includegraphics[width=\widthplot]{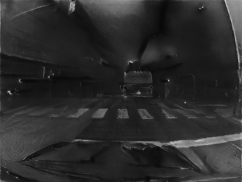} &
	\includegraphics[width=\widthplot]{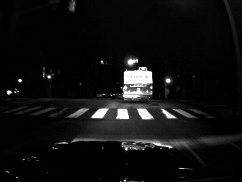} \\

	\rotatebox[origin=l]{90}{night3} &
	\includegraphics[width=\widthplot]{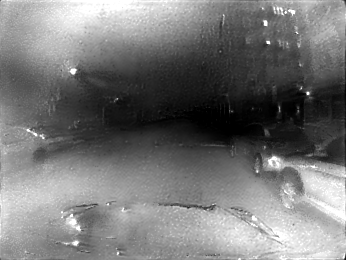} &
	\includegraphics[width=\widthplot]{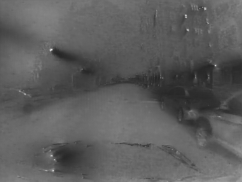} &
	\includegraphics[width=\widthplot]{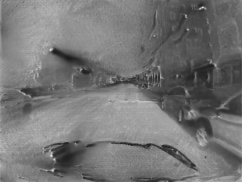} &
	\includegraphics[width=\widthplot]{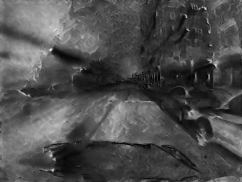} &
	\includegraphics[width=\widthplot]{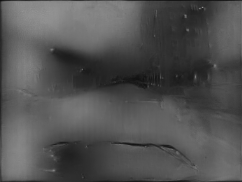} &
	\includegraphics[width=\widthplot]{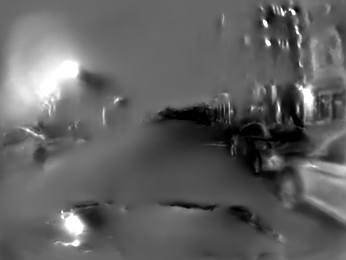} &
	\includegraphics[width=\widthplot]{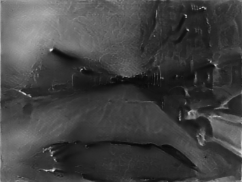} &
	\includegraphics[width=\widthplot]{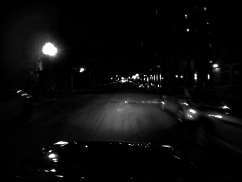} \\
 
	\rotatebox[origin=l]{90}{night3} &
	\includegraphics[width=\widthplot]{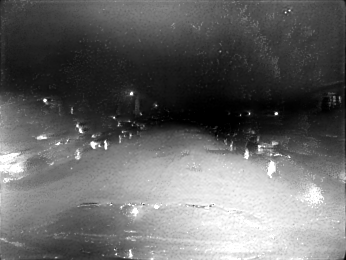} &
	\includegraphics[width=\widthplot]{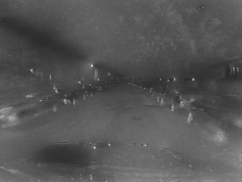} &
	\includegraphics[width=\widthplot]{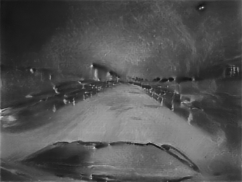} &
	\includegraphics[width=\widthplot]{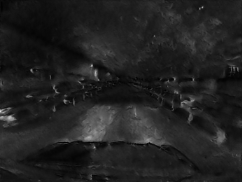} &
	\includegraphics[width=\widthplot]{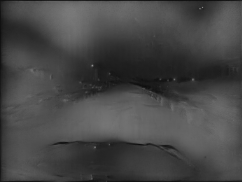} &
	\includegraphics[width=\widthplot]{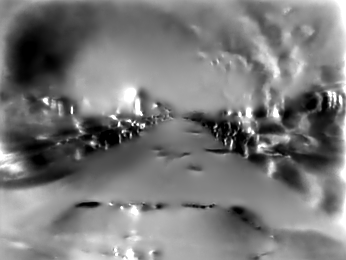} &
	\includegraphics[width=\widthplot]{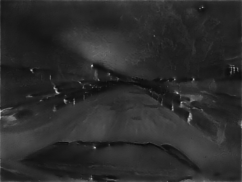} &
	\includegraphics[width=\widthplot]{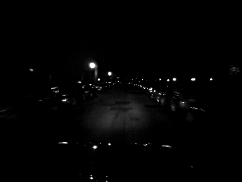} \\

	& E2VID & FireNet & E2VID+ & FireNet+ & \makebox[0pt][c]{SPADE-E2VID} & SSL-E2VID & ET-Net & Reference 

\end{tabular}
	\caption{Additional qualitative comparisons on the night sequences of the MVSEC dataset (MVSEC-NIGHT).}
	\label{fig:qual_eval_MVSEC_night}
        \vspace{1.5cm}
\end{figure*}

Here, we provide qualitative comparisons for various sequences from the ECD, MVSEC, HQF, BS-ERGB, ECD-FAST, and MVSEC-NIGHT datasets. We present these results in Figures~\ref{fig:qual_eval_ECD}-\ref{fig:qual_eval_MVSEC_night}.

\newpage
{\small
\bibliographystyle{ieee_fullname}
\bibliography{supplement}
}